\pdfoutput=1

\documentclass[11pt]{article}

\usepackage[final]{acl}
\usepackage{times}
\usepackage{latexsym}

\usepackage[T1]{fontenc}

\usepackage[utf8]{inputenc}

\usepackage{microtype}

\usepackage{inconsolata}
\usepackage{amsmath}

\usepackage{booktabs}
\usepackage{multirow}
\usepackage{graphicx}
\usepackage{amssymb} 
\usepackage{xcolor}
\usepackage{subfigure}
\usepackage{algorithm} 
\usepackage{algorithmic}
\floatname{algorithm}{Algorithm}

\newcommand{\eg}{\textit{e}.\textit{g}.}
\definecolor{ForestGreen}{rgb}{0.13, 0.55, 0.13}

\title{\textsc{MInD}: A Multi-agent Framework for Zero-shot Harmful Meme Detection}

\author{
 \textbf{Ziyan Liu},
 \textbf{Chunxiao Fan}\textsuperscript{\textdagger},
 \textbf{Haoran Lou},
 \textbf{Yuexin Wu},
 \textbf{Kaiwei Deng}
 \\
 \text{Beijing University of Posts and Telecommunications}\\
 \texttt{\{liuziyan,cxfan\}@bupt.edu.cn}
}

\begin{document}
\maketitle
\renewcommand{\thefootnote}{\textdagger}
\footnotetext{Corresponding author.}
\renewcommand{\thefootnote}{\arabic{footnote}}
\begin{abstract}

The rapid expansion of memes on social media has highlighted the urgent need for effective approaches to detect harmful content. However, traditional data-driven approaches struggle to detect new memes due to their evolving nature and the lack of up-to-date annotated data. To address this issue, we propose \textsc{MInD}, a multi-agent framework for zero-shot harmful meme detection that does not rely on annotated data.  \textsc{MInD} implements three key strategies: 1) We retrieve similar memes from an unannotated reference set to provide contextual information. 2) We propose a bi-directional insight derivation mechanism to extract a comprehensive understanding of similar memes. 3) We then employ a multi-agent debate mechanism to ensure robust decision-making through reasoned arbitration. Extensive experiments on three meme datasets demonstrate that our proposed framework not only outperforms existing zero-shot approaches but also shows strong generalization across different model architectures and parameter scales, providing a scalable solution for harmful meme detection. The code is available at \url{https://github.com/destroy-lonely/MIND}.
\end{abstract}

\section{Introduction}

With the rapid expansion of social media platforms, a new multimodal entity known as the meme has emerged. A meme typically comprises an image combined with a concise textual element, enabling it to spread swiftly across the internet. Memes have become a prevalent form of multimodal content.  While often intended to be humorous, memes are increasingly created or manipulated to convey harmful messages, particularly when they are used to exploit political and socio-cultural divides. 

\begin{figure}[t]
    \centering
    \includegraphics[width=\linewidth]{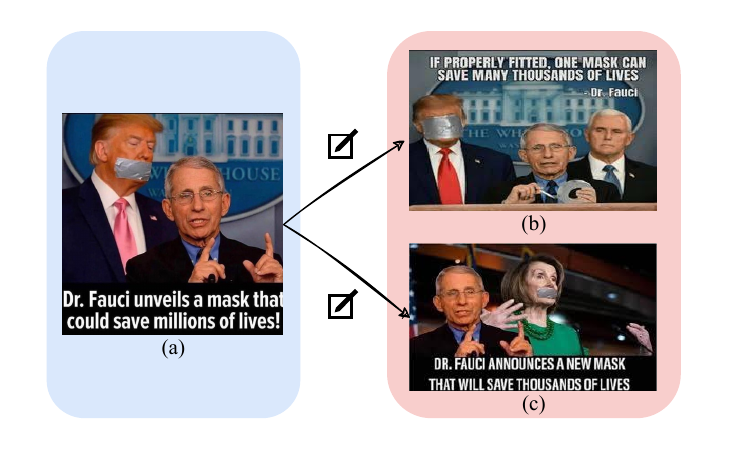}
    \vspace{-0.4cm}
    \caption{Example of trending memes on social media, where (b) and (c) are modified versions of (a).
    \textbf{Meme text}: (a) ``\textit{Dr. Fauci unveils a mask that could save millions of lives!}''; (b) ``\textit{IF PROPERLY FITTED, ONE MASK CAN SAVE MANY THOUSANDS OF LIVES Dr. Fauci}''; (c) ``\textit{DR. FAUCI ANNOUNCES A NEW MASK THAT WILL SAVE THOUSANDS OF LIVES}''}
    \label{fig:intro}
    \vspace{-0.4cm}
\end{figure}

Such memes are referred to as harmful memes\footnote{\color{red}\textbf{Disclaimer:} This paper contains content that may be disturbing to some readers.} and are generally defined as ``multimodal units consisting of an image and embedded text that have the potential to cause harm to an individual, an organization, a community, or society'' ~\cite{meme_survey}. These harmful memes can spread misinformation or perpetuate harmful stereotypes, posing risks to individuals, organizations, and communities. For example, during the COVID-19 pandemic, a frequently shared meme shown in Figure~\ref{fig:intro}(a) featured Dr. Fauci and President Trump at White House press briefings, which was often manipulated and repurposed by various groups skeptical of public health measures. This meme not only undermined both figures' public credibility but also potentially damaged broader public health communication efforts during the pandemic. Therefore,  
automatically understanding and detecting harmful memes becomes increasingly important in maintaining social harmony and integrity on social media.

Previous studies in detecting harmful memes have primarily utilized data-driven multimodal models~\cite{train_detecting, train_multimodal, training_vilio, cao2023prompting, training_towards, lin2023beneath, training_momenta}, which depend on large volumes of high-quality annotated data. These models encounter significant challenges in identifying new memes that quickly emerge in response to current events, as rapidly collecting and annotating sufficient data is challenging. While recent research has explored few-shot in-context learning to enhance detection capabilities with minimal annotations~\cite{modhate, lorehm}, even these approaches still depend on pre-existing annotated data, limiting their adaptability to the fast-paced evolution of harmful memes.

This persistent challenge of data annotation underscores the critical need for methods that can operate without labeled data. We advocate that zero-shot approaches merit more attention, as they are crucial for developing detection systems with the adaptability required to effectively navigate the ever-evolving landscape of memes. Our key insight is that despite evolving into new formats, memes often retain core characteristics that can be identified through careful analysis of similar examples. For instance, as depicted in Figure~\ref{fig:intro}, a meme featuring a White House press briefing could be modified and repurposed in various ways, while maintaining the same core elements from the original setting.

Inspired by this observation and the advanced reasoning capabilities of Large Multimodal Models (LMMs), we propose a novel framework:  \textbf{\textsc{MInD}}, \underline{\textbf{M}}ulti-agent \underline{\textbf{\textsc{In}}}sight derivation for harmful meme \underline{\textbf{D}}etection. Our approach leverages a multi-agent framework that mimics human collaborative analysis: 1) First, it retrieves similar memes from an unannotated reference set to provide contextual information. 2) Then, through bi-directional insight derivation, two agents collaboratively process these memes to extract comprehensive understanding. 3) Finally, multi-agent debate mechanism enables agents to evaluate the derived insights and resolve potential conflicts through reasoned arbitration. 
Through these strategies, our approach enables robust harmful meme detection by leveraging multi-agent reasoning on patterns observed across similar memes, effectively adapting to new and evolving content without relying on annotated data.
Our contributions can be summarized in three folds:
\vspace{-0.2cm}
\begin{itemize}
\item To the best of our knowledge, we pioneer the use of a novel multi-agent framework for zero-shot harmful meme detection, which eliminates the need for annotated data.

\vspace{-0.21cm}

\item We propose \textsc{MInD}, a multi-agent framework that analyzes retrieved similar memes through a novel bidirectional insight derivation and leverages a debate-based reasoning mechanism for robust harm detection.

\vspace{-0.21cm}

\item Extensive experiments on three meme datasets demonstrate that our proposed framework significantly outperforms existing zero-shot state-of-the-art baselines for harmful meme detection.

\end{itemize}

\section{Related work}

\subsection{Harmful Meme Detection}
Harmful meme detection has emerged as a significant research focus, moving forward with the development of extensive benchmarks~\cite{training_momenta, pramanick2021detecting, mami, fhm}. 
Due to the inherently multimodal nature of memes, which incorporate both text and imagery, conventional unimodal approaches~\cite{simonyan2014very, he2016deep, devlin2018bert, raffel2020exploring} have often proven insufficient. In response, recent research has increasingly adopted multimodal strategies~\cite{dosovitskiy2020image, radford2021learning}, aiming to improve detection efficacy by integrating both textual and visual data.

Previous research has leveraged classical two-stream models that combine textual and visual elements, using text and image encoders to capture these features. These systems often employ attention-based mechanisms and multimodal fusion techniques for classifying harmful memes \cite{kiela2019supervised, fhm, suryawanshi2020multimodal, training_momenta}. Another approach involves fine-tuning pre-trained multimodal models to specifically address the task of harmful meme detection \cite{train_multimodal, training_vilio, train_detecting, hee2022explaining}. Additionally, recent efforts have explored various strategies such as data augmentation \cite{zhou2021multimodal, zhu2022multimodal}, ensemble methods \cite{zhu2020enhance, train_detecting, sandulescu2020detecting}, harmful target disentanglement \cite{lee2021disentangling}, and prompt-based tuning \cite{cao2023prompting, ji2023identifying, cao2023pro}. However, most of these approaches rely heavily on large-scale annotated data, which limits their ability to adapt to emerging events and novel harmful content patterns. Although recent studies have explored few-shot in-context learning approaches to address this challenge in low-resource scenarios \cite{modhate, lorehm}, these approaches still fundamentally depend on annotated examples, making them insufficient for detecting harmful memes during emerging events where annotated data is scarce.
In this work, we leverage LMMs to derive insights from similar memes and subsequently employ a multi-agent debate to arrive at comprehensive judgments on meme harmfulness. Without modifying models' weights or requiring annotated data, our approach offers a significant advantage in adapting to real-world scenarios.
\subsection{LLM-Based Multi-Agent Frameworks}
The integration of Large Language Models (LLMs) as agents spans various domains, showcasing their robust planning and reasoning capabilities in diverse settings~\cite{wang2023voyager, yao2022react, shen2023hugginggpt, mu2023embodiedgpt, hong2023metagpt, liu2023agentbench, zhao2024expel, sun2023adaplanner, song2023llm, miao2023selfcheck, madaan2024self}. These advancements underscore the ability of LLMs to tackle complex tasks with minimal supervision. 
Building on the success of single-agent, multi-agent frameworks~\cite{park2023generative, hong2023metagpt, du2023improving, liang2023encouraging, wang2024mobile, qian2024chatdev, tao2024magis, zeng2024autodefense, d2024marg, huang2023agentcoder} facilitate complex interactions and collaborative problem solving, simulating environments where multiple agents work in unison. However, existing frameworks often rely on environmental feedback to iteratively refine their decisions, which becomes impractical for zero-shot binary classification tasks like harmful meme detection, where receiving feedback would effectively reveal the correct answer. In this work, we propose a novel framework that leverages vision-language retrieval enhancement, relevant insight derivation, and multi-agent debate mechanism to enable reliable zero-shot harmful meme detection through collaborative reasoning among specialized agents. 

\begin{figure*}[t]
    \centering
    \includegraphics[width=\linewidth]{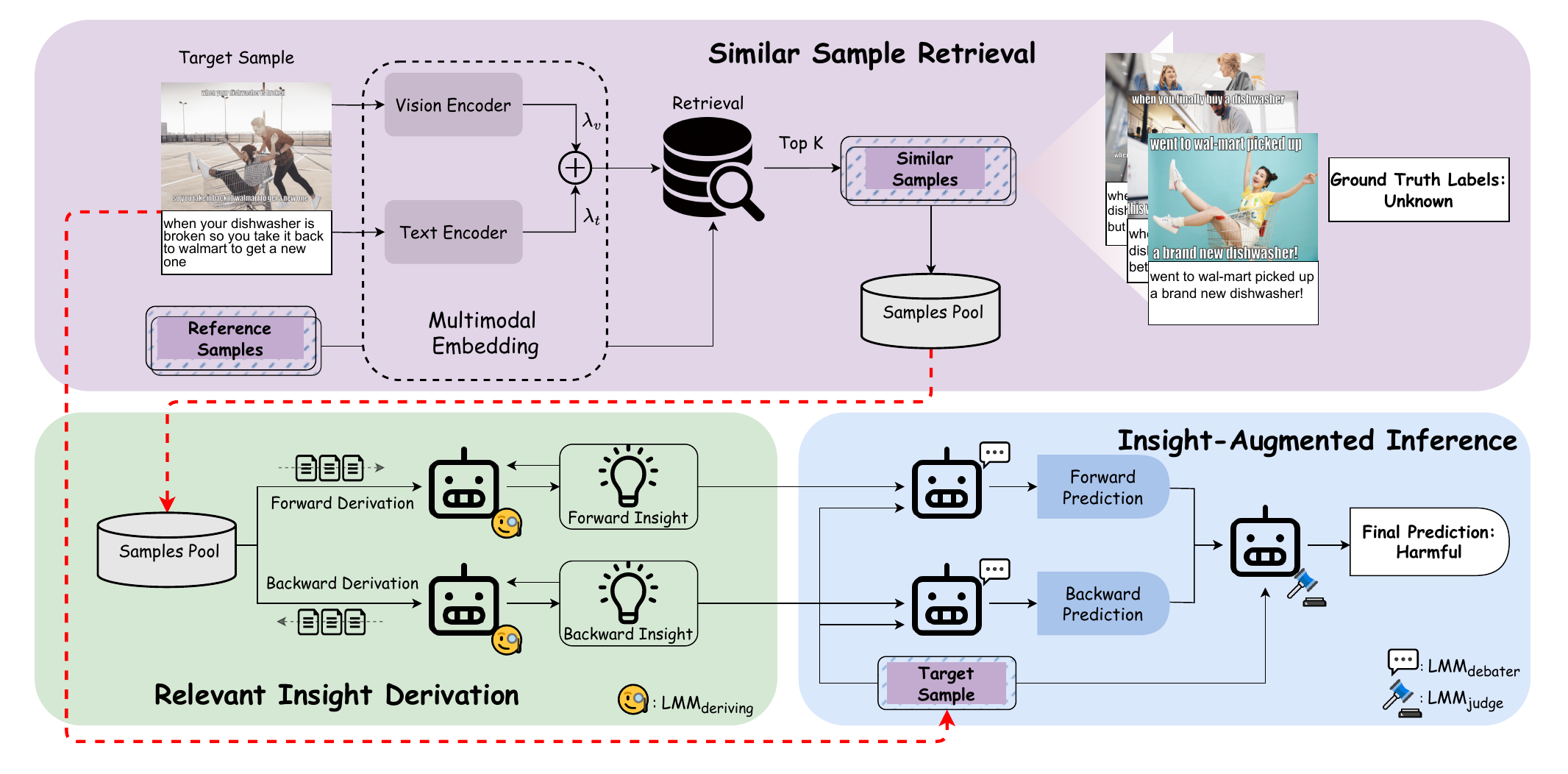}
    \vspace{-0.6cm}
    \caption{An overview of our framework. \textsc{MInD}, for zero-shot harmful meme detection.}
    \label{fig:mind}
    \vspace{-0.4cm}
\end{figure*}

\section{Method}

\subsection{Overview}

\paragraph{Problem Statement}

We define a harmful meme detection dataset as a collection of memes, where each meme $M$ is represented as a tuple $\{\mathcal{V}, \mathcal{T}\}$, consisting of visual component $\mathcal{V}$ and textual component $\mathcal{T}$.
In this work, we explore zero-shot harmful meme detection by leveraging LMMs in a natural language generation paradigm, where both visual and textual inputs are collaboratively processed by LMM agents to determine the harmfulness of a given meme.

The scarcity of high-quality labeled data has become a critical challenge in harmful meme detection, particularly given the rapid evolution and emergence of new memes~\cite{meme_survey}. To address this challenge, we propose \textsc{MInD}, a multi-agent framework that utilizes a training set $S_{\text{train}}$, accessing only the images and text content without using their true labels to detect harmful memes in a test set $S_{\text{test}}$. Since our method is gradient-free, we denote $S_{\text{train}}$ as the reference set $S_{\text{ref}}$. This approach enables robust detection in real-world scenarios where labeled data is scarce or unavailable.

Our proposed framework coordinates multiple LMM agents to collaboratively analyze memes through three interconnected stages: 1) Similar Sample Retrieval(\S\ref{SSR}), which discovers contextually relevant memes from the unlabeled reference set $S_{\text{ref}}$, 2) Relevant Insight Derivation(\S\ref{RID}), where agents engage in forward and backward reasoning to derive relevant insight from similar memes, and 3) Insight-Augmented Inference(\S\ref{IAI}), where specialized debater and judge agents deliberate to reach a well-reasoned conclusion about memes' harmfulness. Through this multi-agent collaboration, \textsc{MInD} achieves robust harmful meme detection without relying on annotated data. The overview of our framework is shown in Figure~\ref{fig:mind}.
\subsection{Similar Sample Retrieval}

\label{SSR}

Internet memes continuously evolve, yet they often display common underlying patterns~\cite{meme_survey, qu2023evolution}. Inspired by retrieval-augmented generation~\cite{guu2020retrieval, lewis2020retrieval}, we focus on retrieving memes similar to the target one, which enables us to utilize these similar memes as a source of insights, providing context and understanding that enhance our assessment of a meme’s harmfulness.

For a meme sample \(M = \{\mathcal{V}, \mathcal{T}\}\), we derive a multimodal embedding by integrating visual and textual features:
\begin{equation}
\label{encode}
\mathbf{E} = \lambda_v \cdot \mathbf{V_{enc}}(\mathcal{V}) + \lambda_t \cdot \mathbf{T_{enc}}(\mathcal{T}),
\end{equation}
\noindent where $\mathbf{E}$ is the multimodal embedding of $M$, $\mathbf{V_{enc}}(\cdot)$ and $\mathbf{T_{enc}}(\cdot)$ are encoders that extract visual and textual features, respectively. The coefficients $\lambda_v$ and $\lambda_t$ are fixed weights for combining the visual and textual modalities. We then perform the operation as in Equation~\ref{encode} for all meme samples in both $S_{\text{test}}$ and $S_{\text{ref}}$ to obtain the embeddings of all meme samples.

To retrieve the most similar samples, we use cosine similarity to match the multimodal embeddings of the target sample and reference samples. The similarity score \(s\) is computed as follows:
\begin{equation}
s = \text{sim}(\mathbf{E}_{\text{target}}, \mathbf{E}_{\text{ref}}),
\end{equation}
\begin{equation}
\text{sim}(\mathbf{E}_{\text{target}}, \mathbf{E}_{\text{ref}}) = \frac{\mathbf{E}_{\text{target}} \cdot \mathbf{E}_{\text{ref}}}{\|\mathbf{E}_{\text{target}}\| \|\mathbf{E}_{\text{ref}}\|},
\end{equation}

\noindent where $\mathbf{E}_{\text{target}}$ and $\mathbf{E}_{\text{ref}}$ are the embedding for the target sample $M_\text{target}$ and a reference sample $M_\text{ref}$, respectively. \(\text{sim}(\cdot)\) denotes the cosine similarity, and $\|\cdot\|$ denotes the vector norm. All the similarity scores of the candidate memes in the reference set to the target meme could form the similarity vector $S = \{s \mid M_{\text{ref}} \in S_\text{ref}\} \in \mathbb{R}^N$.
Then, the top \(K\) reference samples with the highest similarity scores are selected as follows:
\begin{equation}
M_\text{similar} = \{ M_{\text{ref}} \mid s \in \text{Top}_K(S) \},
\end{equation}
\noindent where \(\text{Top}_K(\cdot)\) is the operation of selecting the \(K\) highest values from the similarity vector $S$. The similar set $M_\text{similar}$ contains only the \(K\) most relevant visual and textual components of the similar reference memes, stored in the samples pool, while maintaining the unannotated nature of the reference set for subsequent analysis. 

\subsection{Relevant Insight Derivation}

\label{RID}

The \(M_\text{similar}\), obtained through the Similar Sample Retrieval process described in Section~\ref{SSR}, offers valuable context that can assist in detecting harmful memes. However, leveraging these similar memes directly can be challenging since \(M_\text{similar}\) lacks explicit ground truth labels. Without proper handling, they might confuse rather than assist the LMMs in making accurate judgments. To address this issue, we introduce a novel agentic approach designed to derive relevant insights from \(M_\text{similar}\), ensuring that the information contributes constructively to the harmful meme detection task.

\subsubsection{Forward Insight Derivation}
For each meme in $M_\text{similar}$, the deriving LMM agent processes these memes sequentially. In each iteration $i$, one meme is input into the agent along with previously derived insights, and a new insight set is generated. This insight set serves two purposes: it captures the understanding of the current meme and provides context for analyzing subsequent memes. This iterative process is formulated as:
\begin{equation}
\resizebox{0.89\hsize}{!}{$
\mathcal{I}_{\text{fwd}, i} = \text{LMM}_\text{deriving}(M_{\text{similar}, i}, \mathcal{I}_{\text{fwd}, i-1}, \mathcal{P}_{\text{deriving}}),
$}
\end{equation}
\noindent where \(\mathcal{I}_{\text{fwd}, i}\) and \(\mathcal{I}_{\text{fwd}, i-1}\) represent the derived insight sets at current and previous iterations respectively, \(M_{\text{similar}, i}\) is the i-th meme from the similar meme set, and $\text{LMM}_\text{deriving}$ is an LMM agent equipped with Chain-of-Thought insight derivation prompt $\mathcal{P}_{\text{deriving}}$ as detailed in Appendix \S\ref{sec:ID} and Figure~\ref{fig:prompt}.

\begin{table*}[!t]
    \centering
\resizebox{0.95\textwidth}{!}{\begin{tabular}{@{}l||cc|cc|cc@{}}
\toprule
Dataset         & \multicolumn{2}{c|}{HarM}                  & \multicolumn{2}{c|}{FHM}                        & \multicolumn{2}{c}{MAMI}                     \\ \midrule
Model           & Accuracy                 & Macro-$\emph{F}_1$                & Accuracy                 & Macro-$\emph{F}_1$                     & Accuracy                 & Macro-$\emph{F}_1$               \\ \midrule \midrule
GPT-4o~\cite{achiam2023gpt}& 67.51& 60.29& 68.80& \multicolumn{1}{c|}{68.25} &  81.00& 81.00\\
Gemini-1.5-Flash~\cite{team2024gemini}&  66.10& 64.18& 60.20& \multicolumn{1}{c|}{58.90} & 76.40& 74.29\\ \midrule
LLaVA-1.5-7B~~\cite{liu2024improved}& 59.23& 49.44& 53.80& \multicolumn{1}{c|}{45.51} & 52.90& 41.53\\
 InstructBLIP-7B~\cite{instructblip}& 51.13& 50.99& 52.00& 48.85& 53.10&46.93\\
 MiniGPT-v2-7B~\cite{chen2023minigpt}& 60.12& 52.39& 51.30& 47.88& 57.40&52.22\\
 OpenFlamingo-9B~\cite{awadalla2023openflamingo}& 63.42& 54.36& 50.50& 49.52& 54.70&49.88\\
LLaVA-1.5-13B~~\cite{liu2024improved}& 62.28& 50.45&  55.20& \multicolumn{1}{c|}{53.01} & 60.10& 55.52\\
 InstructBLIP-13B~\cite{instructblip}& 64.92& 49.61& 55.40& 51.89& 60.00&57.97\\
LLaVA-1.6-34B~\cite{liu2024improved}&                 \underline{67.51}&                  \underline{61.59}&                  \textbf{64.00}& \multicolumn{1}{c|}{\textbf{63.51}} &                 \textbf{71.30}&                      \textbf{71.28}\\  \midrule
\textsc{MInD} (LLaVA-1.5-13B)& \multicolumn{1}{c}{\textbf{68.93}} & \multicolumn{1}{c|}{\textbf{65.19}} & \multicolumn{1}{c}{\underline{60.80}} & \multicolumn{1}{c|}{\underline{60.71}} & \multicolumn{1}{c}{\underline{68.90}} & \underline{68.84}\\\bottomrule
\end{tabular}}
\vspace{-0.2cm}
    \caption{Zero-shot harmful meme detection results on three datasets. The accuracy and macro-averaged F1 scores (\%) are reported as the metrics. All baseline models are equipped with Chain-of-Thought prompt. The best and second best results in open-source setting are in \textbf{bold} and \underline{underlined}, respectively.}
    \label{tab:main_results}
    \vspace{-0.1cm}
\end{table*}

\begin{table*}[!t]
    \centering
    \setlength{\tabcolsep}{0.8mm} 
    \renewcommand\arraystretch{1.25}
    \resizebox{\linewidth}{!}{
        \begin{tabular}{l|ll|ll|ll|ll|ll|ll}
            \toprule
            \multicolumn{1}{l|}{\multirow{3}{*}{Model}} & \multicolumn{4}{c|}{HarM} & \multicolumn{4}{c|}{FHM} & \multicolumn{4}{c}{MAMI} \\ 

            \multicolumn{1}{c|}{} & \multicolumn{2}{c}{\hspace{-1em}Accuracy} & \multicolumn{2}{c|}{\hspace{-1em}Macro-$\emph{F}_1$} & \multicolumn{2}{c}{\hspace{-1em}Accuracy} & \multicolumn{2}{c|}{\hspace{-1em}Macro-$\emph{F}_1$} & \multicolumn{2}{c}{\hspace{-1em}Accuracy} & \multicolumn{2}{c}{\hspace{-1em}Macro-$\emph{F}_1$} \\ \cmidrule[0.5pt](l){2-5}\cmidrule[0.5pt](l){6-9}\cmidrule[0.5pt](l){10-13}  

            \multicolumn{1}{c|}{} & \multicolumn{1}{c}{ori.} & \multicolumn{1}{l}{\hspace{2.0em}\textsc{MInD}} & \multicolumn{1}{c}{ori.} & \multicolumn{1}{l|}{\hspace{1.7em}\textsc{MInD}} & \multicolumn{1}{c}{ori.} & \multicolumn{1}{l}{\hspace{2.0em}\textsc{MInD}} & \multicolumn{1}{c}{ori.} & \multicolumn{1}{l|}{\hspace{1.6em}\textsc{MInD}} & \multicolumn{1}{c}{ori.} & \multicolumn{1}{l}{\hspace{2.0em}\textsc{MInD}} & \multicolumn{1}{c}{ori.} & \multicolumn{1}{l}{\hspace{1.6em}\textsc{MInD}} \\ \midrule

            \multicolumn{1}{l|}{LLaVA-1.5-7B} & 59.23 & \hspace{1.0em}62.71 {\color{ForestGreen} \textsubscript{\textbf{(+3.48)}}} & 49.44 & \hspace{1.0em}57.22 {\color{ForestGreen} \textsubscript{\textbf{(+7.78)}}} & 53.80 & \hspace{1.0em}54.00 {\color{ForestGreen} \textsubscript{\textbf{(+0.20)}}} & 45.51  & \hspace{1.0em}48.28 {\color{ForestGreen} \textsubscript{\textbf{(+2.77)}}} & 52.90 & \hspace{1.0em}53.90 {\color{ForestGreen} \textsubscript{\textbf{(+1.00)}}} & 41.53 & \hspace{1.0em}45.45 {\color{ForestGreen} \textsubscript{\textbf{(+3.92)}}}\\ 

            \multicolumn{1}{l|}{LLaVA-1.5-13B} & 62.28 & \hspace{1.0em}68.93 {\color{ForestGreen} \textsubscript{\textbf{(+6.65)}}} & 50.45 & \hspace{1.0em}65.19 {\color{ForestGreen} \textsubscript{\textbf{(+14.74)}}} & 55.20 & \hspace{1.0em}60.80 {\color{ForestGreen} \textsubscript{\textbf{(+5.60)}}} & 53.01  & \hspace{1.0em}60.71 {\color{ForestGreen} \textsubscript{\textbf{(+7.70)}}} & 60.10 & \hspace{1.0em}68.90 {\color{ForestGreen} \textsubscript{\textbf{(+8.80)}}} & 55.52 & \hspace{1.0em}68.84 {\color{ForestGreen} \textsubscript{\textbf{(+13.32)}}}\\ 

            \multicolumn{1}{l|}{LLaVA-1.6-34B} & 67.51 & \hspace{1.0em}69.49 {\color{ForestGreen} \textsubscript{\textbf{(+1.98)}}} & 61.59 & \hspace{1.0em}66.12 {\color{ForestGreen} \textsubscript{\textbf{(+4.53)}}} & 64.00 & \hspace{1.0em}66.40 {\color{ForestGreen} \textsubscript{\textbf{(+2.40)}}} & 63.51 & \hspace{1.0em}68.38 {\color{ForestGreen} \textsubscript{\textbf{(+4.87)}}} & 71.30 & \hspace{1.0em}73.60 {\color{ForestGreen} \textsubscript{\textbf{(+2.30)}}} & 71.28 & \hspace{1.0em}75.38 {\color{ForestGreen} \textsubscript{\textbf{(+4.10)}}}\\ 

            \multicolumn{1}{l|}{Gemini-1.5-Flash} & 66.10 & \hspace{1.0em}68.36 {\color{ForestGreen} \textsubscript{\textbf{(+2.26)}}} & 64.18 & \hspace{1.0em}66.92 {\color{ForestGreen} \textsubscript{\textbf{(+2.74)}}} & 60.20 & \hspace{1.0em}63.80 {\color{ForestGreen} \textsubscript{\textbf{(+3.60)}}} & 58.90 & \hspace{1.0em}62.50 {\color{ForestGreen} \textsubscript{\textbf{(+3.60)}}} & 76.40 & \hspace{1.0em}78.00 {\color{ForestGreen} \textsubscript{\textbf{(+1.60)}}} & 74.29 & \hspace{1.0em}77.89 {\color{ForestGreen} \textsubscript{\textbf{(+3.60)}}}\\ 

            \bottomrule
        \end{tabular}
    }
    \vspace{-0.2cm}
    \caption{Performance improvements of our proposed framework across different model scales and datasets for zero-shot harmful meme detection. Numbers in {\color{ForestGreen}green} indicate absolute improvements over original models.}
    \label{tab:improvement}
    \vspace{-0.4cm}
\end{table*}

\subsubsection{Backward Insight Derivation}
A challenge arises from the sequential nature of Forward Insight Derivation—earlier memes in the sequence benefit more from accumulated insights, while later memes might be less thoroughly analyzed. To address this imbalance, we propose an additional Backward Insight Derivation round. While completing the forward pass, we process similar memes in reverse order:
\begin{equation}
\resizebox{0.89\hsize}{!}{$
\mathcal{I}_{\text{back}, i} = \text{LMM}_\text{deriving}(M_{\text{similar}, K+1-i}, \mathcal{I}_{\text{back}, i-1}, \mathcal{P}_{\text{deriving}}),
$}
\end{equation}
\noindent where \(\mathcal{I}_{\text{back}, i}\) and \(\mathcal{I}_{\text{back}, i-1}\) represent the derived insight sets at current and previous iterations in backward pass, with $K$ being the total number of similar memes.
Through these complementary forward and backward passes, we obtain two sets of insights \(\mathcal{I}_{\text{fwd}, K}\) and \(\mathcal{I}_{\text{back}, K}\), ensuring each meme in $M_\text{similar}$ receives attention from both sequential perspectives.

\subsection{Insight-Augmented Inference}
\label{IAI}
To ensure robust decision-making, we employ two debater agents leveraging complementary insights derived from forward and backward analysis in Section~\ref{RID}. Each debater processes their respective insight sets along with the target meme to generate reasoning-based judgments:
\begin{equation}
\resizebox{0.85\hsize}{!}{$
\mathcal{J}_{\text{fwd}} = \text{LMM}_{\text{debater}}(\mathcal{I}_{\text{fwd}, K}, \mathcal{V}_{\text{target}}, \mathcal{T}_{\text{target}}),
$}
\end{equation}
\vspace{-0.4cm}
\begin{equation}
\resizebox{0.85\hsize}{!}{$
\mathcal{J}_{\text{back}} = \text{LMM}_{\text{debater}}(\mathcal{I}_{\text{back}, K}, \mathcal{V}_{\text{target}}, \mathcal{T}_{\text{target}}),
$}
\end{equation}
\noindent where \(\text{LMM}_{\text{debater}}\) is an LMM agent that generates judgments based on the derived insights and target meme content, \(\mathcal{J}_{\text{fwd}}\) and \(\mathcal{J}_{\text{back}}\) represent the judgments from forward and backward debater agents, and $\mathcal{V}_{\text{target}}$ and $\mathcal{T}_{\text{target}}$ denote the visual and textual components of the target meme. Each judgment contains both reasoning process and final decision. While the reasoning processes may differ, the final decisions are binary indicators of harmfulness.

When examining the judgments from two debater agents, our decision-making process follows two paths: for consensus cases, we directly adopt their shared judgment; for disagreements, a judge agent arbitrates by analyzing both debaters' reasoning. This process can be formalized as:
\begin{equation}
\resizebox{0.95\hsize}{!}{$
\mathcal{J}_\text{final} = 
\begin{cases} 
\mathcal{J}_{\text{fwd}} & \text{if } \mathcal{J}_{\text{fwd}} = \mathcal{J}_{\text{back}} \\[1ex]
\begin{aligned}
\text{LMM}_\text{judge}(&\mathcal{J}_{\text{fwd}}, \mathcal{J}_{\text{back}}, \\
&\mathcal{V}_{\text{target}}, \mathcal{T}_{\text{target}})
\end{aligned}
& \text{if } \mathcal{J}_{\text{fwd}} \neq \mathcal{J}_{\text{back}}
\end{cases}, 
$}
\end{equation}
\noindent where \(\text{LMM}_\text{judge}\) weighs the competing arguments to reach a final judgment \(\mathcal{J}_\text{final}\). This multi-agent debate mechanism enhances reliability through independent assessments and reduces potential biases through diverse perspectives.

In our experiments, we set the number of similar memes $K$ to 3 and use LLaVA-1.5-13B~~\cite{liu2024improved} as the backbone model for all LMM agents as it offers an optimal balance between computational efficiency and model performance. We also conduct extensive experiments with other LMMs~\cite{liu2024improved, team2024gemini} of varying model sizes and architectures to demonstrate the generalizability of our framework (see \S\ref{sec:improvements}).

\section{Experiment}

\subsection{Experiment Setup}
\paragraph{Datasets} We use three publicly available meme datasets for evaluation: (1) HarM~\cite{pramanick2021detecting}, (2) FHM~\cite{fhm}, and (3) MAMI~\cite{mami}. HarM consists of memes related to COVID-19. FHM was released by Facebook as part of a challenge to crowd-source multimodal harmful meme detection in hate speech solutions. MAMI contains memes that are predominantly derogatory towards women, exemplifying typical subjects of online vitriol. Different from FHM and MAMI, where each meme was labeled as \textit{harmful} or \textit{harmless}, HarM was originally labeled with three classes: \textit{very harmful}, \textit{partially harmful}, and \textit{harmless}. For a fair comparison, we merge the \textit{very harmful} and \textit{partially harmful} memes into the \textit{harmful} class, following the setting of recent work~\cite{training_momenta, cao2023prompting, lin2023beneath, lorehm}.
\vspace{-0.15cm}
\paragraph{Baselines} We compare \textsc{MInD} with state-of-the-art (SOTA) approaches for zero-shot harmful meme detection: 1) \textbf{GPT-4o}~\cite{achiam2023gpt}; 2) \textbf{Gemini-1.5-Flash}~\cite{team2024gemini}; 3) \textbf{LLaVA-1.5-7B}~~\cite{liu2024improved}; 4) \textbf{InstructBLIP-7B}~\cite{instructblip}; 5) \textbf{MiniGPT-v2-7B}~\cite{chen2023minigpt}; 6) \textbf{OpenFlamingo-9B}~\cite{awadalla2023openflamingo}; 7) \textbf{LLaVA-1.5-13B}~~\cite{liu2024improved}; 8) \textbf{InstructBLIP-13B}~\cite{instructblip}; 9)\textbf{LLaVA-1.6-34B}~\cite{liu2024improved}; 10) \textbf{\textsc{MInD} (*)}: Our proposed multi-agent approach based on LLaVA-1.5-13B. We use the accuracy and macro-averaged F1 (dominant) scores as the evaluation metrics.

The data statistics, baseline descriptions and model implementation are detailed in the Appendix \S\ref{sec:datasets} , \S\ref{sec:baselines}, and \S\ref{sec:ID}, respectively.

\begin{table*}[t]
    \centering
    \resizebox{0.90\textwidth}{!}{%
        \begin{tabular}{@{}l||cc|cc|cc@{}}
        \toprule
        Dataset & \multicolumn{2}{c|}{HarM} & \multicolumn{2}{c|}{FHM} & \multicolumn{2}{c}{MAMI} \\ \midrule
        Model & Accuracy & Macro-$F_{\text{\emph{1}}}$ & Accuracy & Macro-$F_{\text{\emph{1}}}$ & Accuracy & Macro-$F_{\text{\emph{1}}}$ \\ \midrule \midrule
        \textsc{MInD} (LLaVA-1.5-13B) & \multicolumn{1}{c}{68.93} & \multicolumn{1}{c|}{65.19} & \multicolumn{1}{c}{60.80} & \multicolumn{1}{c|}{60.71} & \multicolumn{1}{c}{68.90} & 68.84\\
        \hspace{0.5cm} w/o SSR & \multicolumn{1}{c}{64.97} & \multicolumn{1}{c|}{60.92} & \multicolumn{1}{c}{60.40} & \multicolumn{1}{c|}{60.38} & \multicolumn{1}{c}{66.70} & 66.38\\
        \hspace{0.5cm} w/o RID & \multicolumn{1}{c}{62.67} & \multicolumn{1}{c|}{51.93} & \multicolumn{1}{c}{57.20} & \multicolumn{1}{c|}{56.02} & \multicolumn{1}{c}{59.70} & 56.51\\
        \hspace{0.5cm} w/o RID$_\text{forward}$ & \multicolumn{1}{c}{64.97} & \multicolumn{1}{c|}{63.46} & \multicolumn{1}{c}{60.20} & \multicolumn{1}{c|}{59.81} & \multicolumn{1}{c}{66.60} & 66.60\\
        \hspace{0.5cm} w/o RID$_\text{backward}$ & \multicolumn{1}{c}{64.41} & \multicolumn{1}{c|}{62.28} & \multicolumn{1}{c}{59.20} & \multicolumn{1}{c|}{58.94} & \multicolumn{1}{c}{68.00} & 67.98\\
 \hspace{0.5cm} w/o IAI& 63.28& 60.97& 59.00& 58.53& 68.10&68.10\\
 \bottomrule
        \end{tabular}%
    }
    \vspace{-0.1cm}
    \caption{Ablation studies on our proposed framework.}
    \label{tab:ablation}
    \vspace{-0.4cm}
\end{table*}

\subsection{Harmful Meme Detection Performance}
Table~\ref{tab:main_results} illustrates the performance of our proposed framework \textsc{MInD} versus all the compared baselines for zero-shot harmful meme detection. It is observed that:
1) In the first group of closed-source models, GPT-4o and Gemini-1.5-Flash demonstrate competitive performance across all datasets, with GPT-4o showing generally stronger results, particularly on the MAMI dataset.
2) The second group consists of open-source models with varying parameter sizes. Among them, LLaVA-1.6-34B achieves the best baseline performance in the open-source setting. This superior performance can be attributed to its larger parameter size, as we observe a general trend where models with larger parameters (\eg, LLaVA-1.6-34B) outperform their smaller counterparts (\eg, LLaVA-1.5-7B) in this challenging zero-shot task.
3) Our proposed \textsc{MInD} framework, built upon LLaVA-1.5-13B, demonstrates remarkable improvements. Compared to the base LLaVA-1.5-13B model, \textsc{MInD} improves the macro-averaged-F1 scores by 14.74\%, 7.70\%, and 13.32\% on HarM, FHM, and MAMI respectively. Notably, on the HarM dataset, \textsc{MInD}'s performance exceeds LLaVA-34B by 3.60\% and surpasses the closed-source GPT-4o by 4.90\%. Additionally, \textsc{MInD} achieves comparable performance with Gemini-1.5-Flash on FHM, despite being based on a much smaller open-source model. These results demonstrate the effectiveness of our approach even compared to powerful proprietary models, validating the strength of our proposed multi-agent framework in zero-shot harmful meme detection.
\vspace{-0.2cm}
\subsection{Improvements Across Model Scales}
\label{sec:improvements}
To further validate the effectiveness and generalization ability of our \textsc{MInD} framework, we apply it to various LMMs with different parameter sizes, including both open-source and closed-source models. As shown in Table~\ref{tab:improvement}, \textsc{MInD} consistently brings improvements across all models and datasets. For open-source models, we observe that: 1) The improvement is most significant on LLaVA-1.5-13B, with macro-averaged-F1 scores increasing by 14.74\%, 7.70\%, and 13.32\% on HarM, FHM, and MAMI respectively; 2) Even for the stronger LLaVA-1.6-34B model, \textsc{MInD} still achieves notable gains of 4.53\%, 4.87\%, and 4.10\% across the three datasets. Remarkably, \textsc{MInD} also enhances the performance of closed-source model Gemini-1.5-Flash, improving its performance by 2.74\%, 3.60\%, and 3.60\% on the three datasets. Most notably, as shown in Tables~\ref{tab:main_results} and~\ref{tab:improvement}, with our framework, LLaVA-1.5-7B approaches the performance of base LLaVA-1.5-13B, enhanced LLaVA-1.5-13B matches or surpasses base LLaVA-1.6-34B, and both enhanced LLaVA-1.6-34B and Gemini-1.5-Flash achieve competitive performance with GPT-4o, demonstrating the strong capability of our framework in boosting model performance regardless of model scales and architectures.

\subsection{Ablation Study}
To thoroughly evaluate the effectiveness of different strategies in our framework, we conduct ablation studies with several variants of \textsc{MInD}. As shown in Table~\ref{tab:ablation}, we examine five variants: 1) \textit{w/o SSR}: replacing Similar Sample Retrieval (SSR) with random selection of three memes from $S_{\text{ref}}$ as similar references while keeping other strategies unchanged; 2) \textit{w/o RID}: removing Relevant Insight Derivation (RID), which consequently eliminates Insight-Augmented Inference (IAI) as it relies on derived insights, leaving similar memes to be directly used for reasoning; 3) \textit{w/o RID}$_\text{forward}$: removing forward insight derivation and its corresponding debate in IAI, where only backward insights are used for reasoning; 4) \textit{w/o RID}$_\text{backward}$: removing backward insight derivation and its corresponding debate in IAI, where only forward insights are used for reasoning; 5) \textit{w/o IAI}: removing the multi-agent debate mechanism, where both forward and backward insights are directly used together for reasoning.

The ablation results reveal several interesting findings: 1) Random selection of similar memes (\textit{w/o SSR}) leads to significant drops in macro-averaged-F1 scores (4.27\%, 0.33\%, and 2.46\% on three datasets), suggesting that similarity-based retrieval is crucial for finding relevant reference memes; 2) Direct use of similar memes without insight derivation (\textit{w/o RID}) causes the most substantial degradation in F1 scores (13.26\%, 4.69\%, and 12.33\%), indicating that raw similar memes might introduce noise without proper insight derivation; 3) Using single-direction insight derivation (\textit{w/o RID}$_\text{forward}$ or \textit{w/o RID}$_\text{backward}$) results in moderate performance drops, with \textit{w/o RID}$_\text{forward}$ showing slightly better performance than \textit{w/o RID}$_\text{backward}$ on HarM and FHM datasets, while \textit{w/o RID}$_\text{backward}$ performs marginally better on MAMI; 4) Without the multi-agent debate mechanism (\textit{w/o IAI}), the F1 scores decrease by 4.22\%, 2.18\%, and 0.74\%, demonstrating that the multi-agent debate mechanism helps reconcile potentially conflicting insights. 
These results highlight that all strategies in our framework, Similar Sample Retrieval, Relevant Insight Derivation, and Insight-Augmented Inference, play essential and complementary roles in harmful meme detection. SSR retrieves similar memes as references, RID builds upon these retrieved memes to derive bidirectional insights through forward and backward reasoning, and IAI leverages the derived insights to generate robust judgments through multi-agent debate. The synergy among these strategies significantly enhances the framework's robustness in detecting harmful content, as evidenced by the performance degradation when any strategy is disabled.

\begin{figure}[t]
    \centering
    \includegraphics[width=1\linewidth]{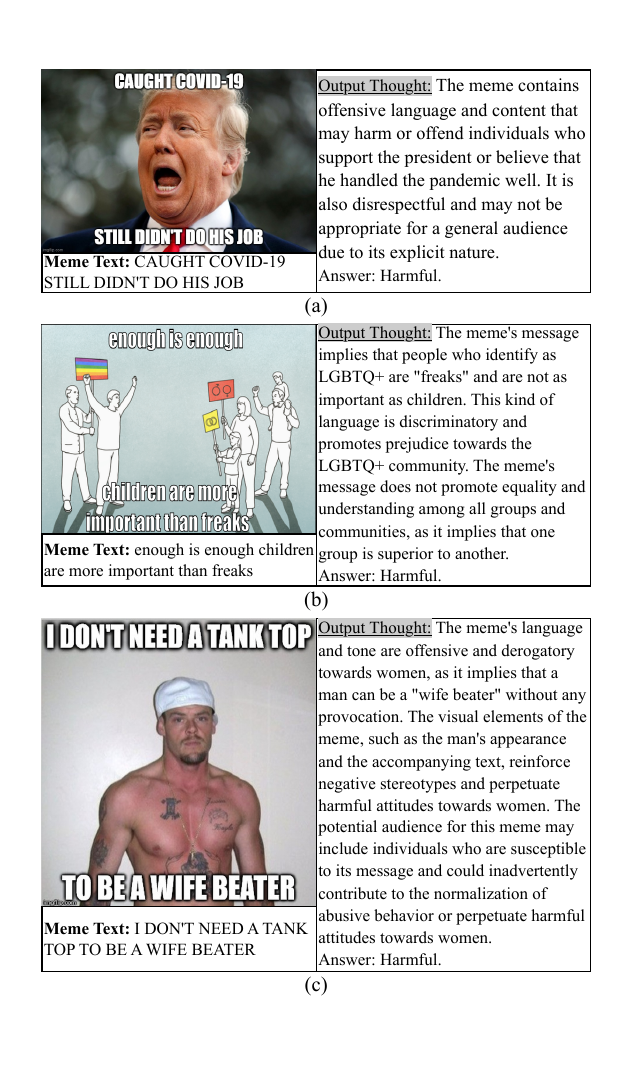}
    \vspace{-0.8cm}
    \caption{Examples of correctly predicted harmful memes in (a) HarM, (b) FHM,  and (c) MAMI datasets. }
    \label{fig:case_study}
    \vspace{-0.4cm}
\end{figure}

\begin{figure}[t]
    \centering
    \includegraphics[width=1\linewidth]{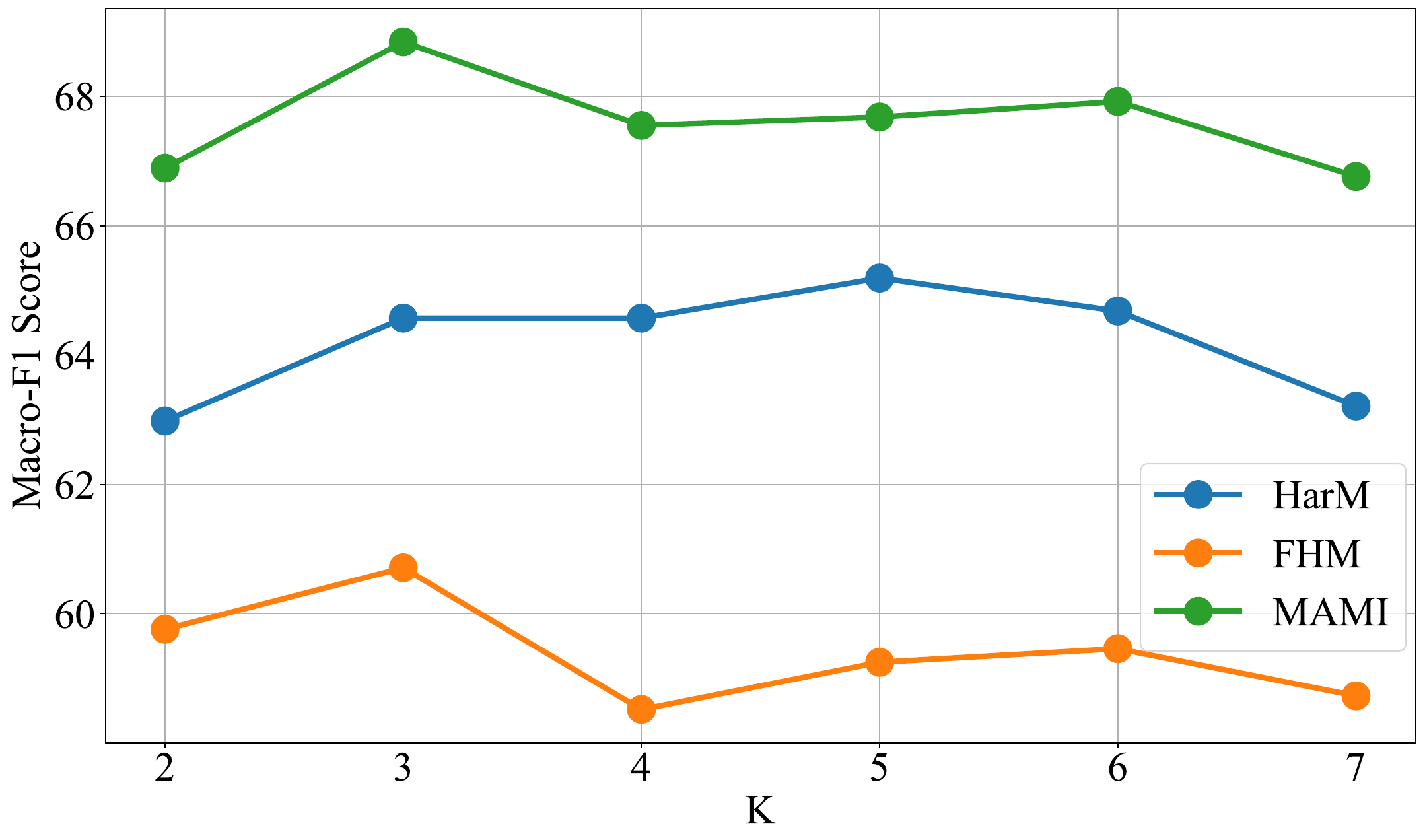}
    \vspace{-0.8cm}
    \caption{Effect of $Top_K$ in Similar Sample Retrieval}
    \label{fig:topk}
    \vspace{-0.4cm}
\end{figure}

\subsection{Effect of Retrieved Meme Count}
To investigate the impact of similar memes in our framework, we conduct experiments examining performance variations with different numbers of retrieved memes \(K\) as shown in Figure~\ref{fig:topk}. We observe that: 1) All three datasets show relatively consistent performance trends, with peak macro-averaged-F1 scores generally occurring at lower \(K\) values before gradually declining. However, as the \(K\) value further increases, this score gradually decreases. This indicates that there exists an optimal number of similar memes that can be effectively utilized by the framework to enhance its discriminative ability. 2) Larger \(K\) values don't necessarily lead to better results, as they tend to incorporate less similar memes into the retrieval results, potentially introducing noise rather than beneficial information.  
Based on these observations, setting \(K\) to 3 achieves the optimal balance between performance and efficiency across all datasets.

\subsection{Case Study}
To better understand how our proposed framework processes and evaluates memes, we analyze several correctly predicted cases, where we show important content in the thought, as shown in Figure~\ref{fig:case_study}. 

From analyzing the output thought, we observe that:
1) Our proposed framework effectively connects multimodal information between meme text and imagery using commonsense knowledge. For example, in Figure~\ref{fig:case_study}(a), it recognizes how ``caught COVID-19'' and ``didn't do his job'' in the text directly relates to the presidential image, forming a critical commentary. In Figure~\ref{fig:case_study}(b), the framework links the protest imagery with the text's discriminatory comparison between ``children'' and ``freaks'', identifying its target toward the LGBTQ+ community shown in the visual elements. For Figure~\ref{fig:case_study}(c), it connects the aggressive text about being a ``wife beater'' with the threatening visual presentation to recognize harmful gender-based messaging.
2) Through deriving insights from similar memes, our proposed framework demonstrates sophisticated analysis of broader implications. In Figure~\ref{fig:case_study}(a), the reasoning considers how comparable political memes during the pandemic have contributed to increased social division and undermined public health messaging. The analysis in Figure~\ref{fig:case_study}(b) draws parallels with other discriminatory content to show how such messaging creates dangerous hierarchies between social groups and perpetuates harmful stereotypes. For Figure~\ref{fig:case_study}(c), by connecting to similar cases of normalized violence, the reasoning reveals how such content can desensitize viewers to domestic abuse and reinforce dangerous attitudes toward women.
Through these examples, we can see how our framework provides clear, interpretable reasoning that connects visual and textual elements while considering real-world implications. This analytical transparency can be valuable for human checkers verifying model predictions in zero-shot setting. 
Additional case study and error analysis of our proposed framework are provided in Appendix \S\ref{sec:more_cases} and \S\ref{sec:error_analysis}.

\section{Conclusion and Future Work}
\vspace{-0.11cm}
In this paper, we delved into the zero-shot harmful meme detection task that operates without annotated training data. To this end, we proposed \textsc{MInD}, a novel multi-agent framework that seamlessly integrates Similar Sample Retrieval, Relevant Insight Derivation, and Insight-Augmented Inference to enable robust harmful content detection. Through comprehensive experiments and analyses on three meme datasets, we demonstrated the effectiveness of our proposed framework and the importance of each strategy. Future efforts aim to enhance our research by exploring the robustness of relevant insight derivation across diverse meme contexts.

\section*{Limitations}
There are multiple ways to further improve this work:
\begin{itemize}
    \item The framework's performance partially depends on the quality of retrieved similar memes. While our current similarity-based retrieval mechanism shows effectiveness, selecting truly relevant reference memes remains challenging. Current embedding-based similarity metrics may not fully capture the complex semantic relationships of highly contextual or novel meme formats, suggesting that more sophisticated retrieval strategies could enhance performance.
    \item Although our bi-directional insight derivation mechanism enables comprehensive analysis, its uniform treatment of all retrieved memes might not optimally capture their varying degrees of relevance. Implementing a more nuanced weighting mechanism that considers relative importance could lead to more refined insights.
    \item The current multi-agent debate mechanism primarily focuses on achieving consensus through reasoned arbitration. However, quantitatively evaluating the quality and reliability of derived insights remains challenging, particularly when correct conclusions stem from potentially flawed reasoning chains. This limitation makes it difficult to systematically assess the framework's overall robustness.
    \item While \textsc{MInD} eliminates substantial training and data annotation costs, its inference time can be significant due as it requires numerous LMM calls. This results in an approximate 8x computational overhead compared to simple zero-shot baselines, which is a key consideration for real-time deployment. Nevertheless, our ablation studies demonstrate that specific configurations, such as employing unidirectional RID, can halve these calls, offering a flexible efficiency-performance trade-off.
    \item Despite efforts to ensure robust decision-making through multi-agent debate, the framework's performance still heavily relies on the base LMM's understanding of social and cultural nuances. It may struggle to identify subtly encoded harmful content or evolving memes requiring deep cultural knowledge that human moderators readily recognize.
\end{itemize}

\section*{Ethics Statement}
Our research aims to combat harmful meme content through zero-shot detection methods, contributing to safer online spaces. The harmful content types addressed in our study are well-documented concerns in social media research. Our work focuses on detecting various forms of harmful content including hate speech, misogyny, and misinformation that can negatively impact individuals and communities.
However, we are aware of the potential for malicious users to reverse-engineer and create memes that go undetected or misunderstood by AI systems based on \textsc{MInD}. We strongly condemn such practices and emphasize that our research is intended solely for scientific purposes and harmful content prevention. The framework and associated resources are strictly prohibited from commercial use or malicious exploitation.
To ensure responsible development and evaluation of our framework, we implemented several protective measures: 1) all experiments were conducted using publicly available research datasets, following their respective usage agreements; 2) no personal user data was collected or utilized in this study.
We believe the benefits of advancing harmful meme detection capabilities outweigh the potential risks, particularly given the growing challenge of moderating harmful content on social media. The opinions and content contained in the meme samples should not be interpreted as representing the views of the authors. Our framework is designed to assist, not replace, human moderation efforts in maintaining healthy online communities.

\section*{Acknowledgements}
This work is supported in part by the National Natural Science Foundation of China (Grant Nos. 62376034 and 92467105).

\bibliography{custom}
\newpage
\appendix

\begin{table}[t] \small
\centering
\resizebox{0.35\textwidth}{!}{\begin{tabular}{@{}cccc@{}}
\toprule
\multirow{2}{*}{Datasets} & \multicolumn{2}{c}{Test} \\
                          & \#harmful   & \#harmless   \\ \midrule
HarM                    & 124        & 230         \\
FHM                       & 250        & 250         \\
MAMI                    & 500        & 500         \\\bottomrule
\end{tabular}}
\caption{Statistics of test sets.}
\label{tab:statistics}
\end{table}

\section{Datasets}
\label{sec:datasets}
The detailed statistics for the original test splits of the three datasets are shown in Table~\ref{tab:statistics}.

\section{Baselines} 
\label{sec:baselines}
We compare \textsc{MInD} with several state-of-the-art (SOTA) methods for zero-shot harmful meme detection:
\textbf{GPT-4o}~\cite{achiam2023gpt}: a proprietary large multimodal model by OpenAI that demonstrates strong zero-shot capabilities in visual-language tasks through in-context learning;
\textbf{Gemini-1.5-Flash}~\cite{team2024gemini}: Google's latest multimodal model that shows competitive performance in reasoning and visual understanding tasks;
\textbf{LLaVA-1.5-7B}~~\cite{liu2024improved}: a lightweight multimodal model built on Vicuna-7B, trained on diverse visual instruction data for general vision-language tasks;
\textbf{InstructBLIP-7B}~\cite{instructblip}: an instruction-tuned vision-language model based on BLIP-2 architecture that leverages Vicuna-7B for language modeling;
\textbf{MiniGPT-v2-7B}~\cite{chen2023minigpt}: a compact yet effective multimodal model that combines visual encoding with instruction-tuned language generation;
\textbf{OpenFlamingo-9B}~\cite{awadalla2023openflamingo}: an open-source implementation of Flamingo models that enables frozen language models to process visual inputs through cross-attention;
\textbf{LLaVA-1.5-13B}~~\cite{liu2024improved}: a medium-sized variant of LLaVA built on Vicuna-13B, with enhanced visual grounding and reasoning capabilities;
\textbf{InstructBLIP-13B}~\cite{instructblip}: an enhanced version of InstructBLIP using Vicuna-13B as the language model backbone;
\textbf{LLaVA-1.6-34B}~\cite{liu2024improved}: the latest and largest version of LLaVA with improved reasoning, OCR, and world knowledge capabilities;
\textbf{\textsc{MInD} (*)}: our proposed multi-agent approach for zero-shot harmful meme detection,  based on LLaVA-1.5-13B.
We use the accuracy and macro-averaged F1 (dominant) scores as the evaluation metrics, where the macro-averaged F1 score is the more important metric owing to the imbalanced class prevalence, to capture competitive performance beyond the majority class. 

For baseline methods, we implement the following Chain-of-Thought~\cite{kojima2022large} prompt structure, which achieves better performance than direct classification:
``\textit{Given the meme, with the Text: "$\{\mathcal{T}\}$" embedded in the image $\{\mathcal{V}\}$, your task is to determine whether this meme is harmful or not, in order to maintain the harmlessness and integrity of information on the Internet. Please leverage your extensive knowledge to deeply analyze and understand this meme, and give your final judgment.
Your output should stricly follow the format: "Thought: [Your analysis]
Answer: [harmful/harmless].}''

\section{Implementation Details}
\label{sec:ID}
\begin{figure*}[t]
    \centering
    \includegraphics[width=1\linewidth]{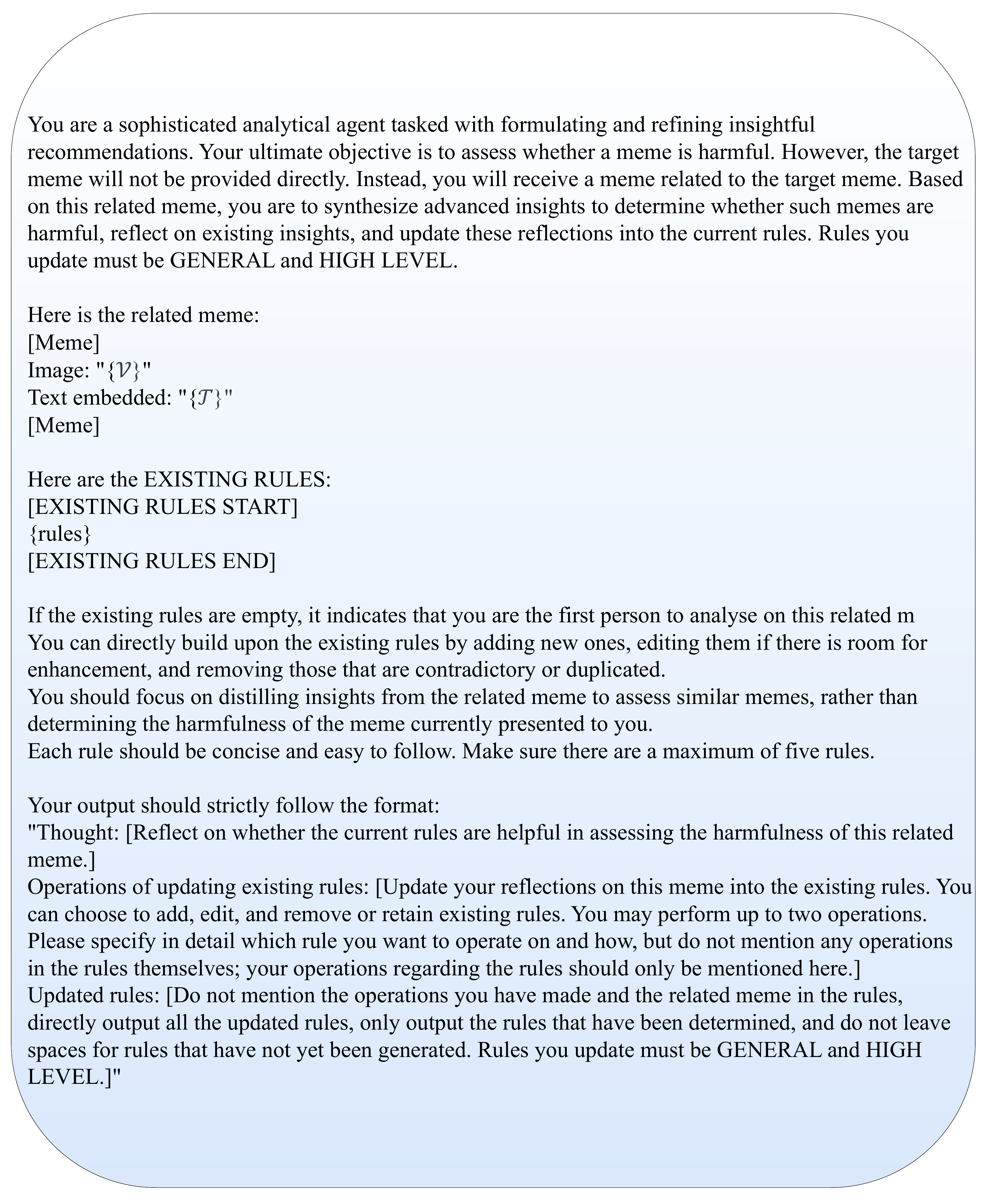}
    \caption{The prompt of $\mathcal{P}_{\text{deriving}}$}
    \label{fig:prompt}
\end{figure*}

For our proposed framework, we conduct extensive experiments with various LMMs including LLaVA series (LLaVA-1.5-7B, LLaVA-1.5-13B, LLaVA-1.6-34B), and Gemini-1.5-Flash as our LMM agents. For the baseline of the closed-source model, we use the “gpt-4o-2024-11-20” and “gemini-1.5-flash” versions, respectively. To ensure reproducibility and consistent comparison, we set the temperature parameter to 0 without any sampling mechanism for all models. For the multimodal embedding computation in Equation~\ref{encode}, we employ CLIP~\cite{radford2021learning} with the specific version ``ViT-L/14@336p'' as our frozen pretrained vision and text Transformer encoders ($\mathbf{V_{enc}}$ and $\mathbf{T_{enc}}$). The modality fusion weights $\lambda_v$ and $\lambda_t$ are set to 0.8 and 0.2 respectively through grid search optimization within [0,1].
In our implementation, we set $K$ to 3 for retrieving the most similar reference samples in Similar Sample Retrieval, which provides a balance between computational efficiency and sufficient context information.
For the Relevant Insight Derivation process, we design the prompt template as Figure~\ref{fig:prompt} to guide the $\text{LMM}_\text{deriving}$ agent.

In the Insight-Augmented Inference process, for the $\text{LMM}_\text{debater}$ agent, we implement the following prompt structure:
``\textit{Given the meme, with the Text: "$\{\mathcal{T}\}$" embedded in the image $\{\mathcal{V}\}$.
Is it harmful or harmless?
Note: [$\{note\}$] 
(Consider this note above with caution - it may contain biases or subjective interpretations that could influence judgment. Evaluate its reliability and relevance carefully)
Your output should strictly follow the format:
"Thought: [First, analyze the meme's content independently. Then, carefully consider how the provided note may inform or bias your understanding. Weigh the note's credibility and relevance before incorporating it into your final assessment.]
Answer: [harmful/harmless]."}''

When disagreement occurs, the judge agent utilizes this prompt template:
``\textit{Given the meme, with the Text: "$\{\mathcal{T}\}$" embedded in the image $\{\mathcal{V}\}$.
Is it harmful or harmless?
Regarding this meme, there are two debaters with differing opinions.
Debater 1 believes that:
The correct answer should be: $\{debater1answer\}$
Debater 1's reason: [$\{debater1reason\}$]
Debater 2 believes that:
The correct answer should be: $\{debater2answer\}$
Debater 2's Reason: [$\{debater2reason\}$]
As a judge, you need to determine who is correct based on the viewpoints of the two debaters and provide the correct answer that you believe to be true.
Your response should strictly adhere to this format:
"Thought: [Based on the meme itself and your careful consideration, whose viewpoint do you think is correct? And why?]
Answer: [Your final judgment(harmful/harmless)].''}

All experiments were conducted using four NVIDIA A40 48GiB GPUs. The total processing time varies by dataset size: approximately  1.5 hours for HarM, 3 hours for FHM, and 5 hours for MAMI when using Gemini-1.5-Flash. With LLaVA-1.5-13B, these times extend to 3 hours, 4.5 hours, and 9 hours respectively.

Due to privacy and ongoing research considerations, the code used in this study is not included in the submission. However, we commit to making the code publicly available upon the acceptance of this paper.

\begin{table*}[t]
    \centering
    \resizebox{0.90\textwidth}{!}{%
        \begin{tabular}{@{}l||cc|cc|cc@{}}
        \toprule
        Dataset & \multicolumn{2}{c|}{HarM} & \multicolumn{2}{c|}{FHM} & \multicolumn{2}{c}{MAMI} \\ \midrule
        Model & Accuracy & Macro-$F_{\text{\emph{1}}}$ & Accuracy & Macro-$F_{\text{\emph{1}}}$ & Accuracy & Macro-$F_{\text{\emph{1}}}$ \\ \midrule \midrule
        LLaVA-1.5-13B& \multicolumn{1}{c}{62.28} & \multicolumn{1}{c|}{50.45} & \multicolumn{1}{c}{55.20} & \multicolumn{1}{c|}{53.01} & \multicolumn{1}{c}{60.10} & 55.52\\
        \hspace{0.5cm} zero-shot w/ SSR & \multicolumn{1}{c}{62.67} & \multicolumn{1}{c|}{51.93} & \multicolumn{1}{c}{57.20} & \multicolumn{1}{c|}{56.02} & \multicolumn{1}{c}{59.70} & 56.51\\
        \hspace{0.5cm} zero-shot w/o SSR& \multicolumn{1}{c}{59.04} & \multicolumn{1}{c|}{47.72} & \multicolumn{1}{c}{53.00} & \multicolumn{1}{c|}{52.19} & \multicolumn{1}{c}{56.20} & 50.73\\
        \hspace{0.5cm} 3-shot w/ SSR & \multicolumn{1}{c}{66.10} & \multicolumn{1}{c|}{59.60} & \multicolumn{1}{c}{60.60} & \multicolumn{1}{c|}{60.36} & \multicolumn{1}{c}{66.70} & 66.66\\
        \hspace{0.5cm} 3-shot w/o SSR& \multicolumn{1}{c}{60.45} & \multicolumn{1}{c|}{56.44} & \multicolumn{1}{c}{56.60} & \multicolumn{1}{c|}{58.42} & \multicolumn{1}{c}{61.00} & 60.92\\
 \hspace{0.5cm} w/ \textsc{MInD}& 68.93& 65.19& 60.80& 60.71& 68.90&68.84\\
 \bottomrule
        \end{tabular}%
    }
    \vspace{-0.1cm}
    \caption{Evaluation results comparing with few-shot methods.}
    \label{tab:vs_fewshot}
    \vspace{-0.3cm}
\end{table*}

\section{Related work about LMMs}
LLMs have recently expanded their capabilities beyond text processing to handle image inputs. Building upon these language models, LMMs have emerged as powerful tools for visual-language understanding. 
The current LMM landscape features both commercial and open-source solutions. Industry leaders like GPT-4o~\cite{achiam2023gpt} and Google's Gemini~\cite{team2024gemini} stand out for their strong zero-shot performance and sophisticated visual reasoning capabilities. These models excel at understanding nuanced visual content, engaging in detailed conversations about images, and providing rich, contextual analysis of visual information.
In parallel with these commercial models, the open-source community has achieved significant breakthroughs. LLaVA~\cite{liu2023visual} and its latest version, LLaVA-1.6-34B~\cite{liu2024improved}, have made substantial progress in matching the capabilities of commercial models while keeping their technology transparent and accessible. These open models~\cite{bai2023qwen, chen2024internvl, wang2023cogvlm, chen2023minigpt, instructblip, awadalla2023openflamingo} use clever training methods, including visual instruction tuning and streamlined fine-tuning approaches, to achieve strong results without requiring massive computational power. In this work, we utilize LLaVA-1.5-13B~~\cite{liu2024improved} as our primary backbone model for all LMM agents. To demonstrate the generalizability of our proposed framework, we also conduct experiments with LLaVA-1.6-34B and Gemini-1.5-Flash, which represent the current state-of-the-art in open-source and commercial LMMs respectively.

\begin{algorithm}
\caption{\textsc{MInD} - Similar Sample Retrieval}
\begin{algorithmic}[0]
\STATE \textbf{Initialize:}
\STATE Modality fusion weights $\lambda_v$, $\lambda_t$;
\STATE Visual Encoder $\mathbf{V_{enc}}(\cdot)$;
\STATE Textual Encoder $\mathbf{T_{enc}}(\cdot)$;
\STATE Reference set $S_{\text{ref}}$, Test set $S_{\text{test}}$;
\STATE Target meme $M_{\text{target}}$;
\STATE Number of similar samples $K$;
\STATE Embedding set $\mathbf{E}_{\text{all}} \leftarrow \emptyset$;
\STATE Similarity scores $S \leftarrow \emptyset$;
\STATE Similar memes set $M_{\text{similar}} \leftarrow \emptyset$;

\STATE \textbf{Embedding Generation:}
\FOR{each meme $M \in \{S_{\text{ref}} \cup S_{\text{test}}\}$}
    \STATE $\mathbf{V} \leftarrow \mathbf{V_{enc}}(\mathcal{V})$
    \STATE $\mathbf{T} \leftarrow \mathbf{T_{enc}}(\mathcal{T})$
    \STATE $\mathbf{E} \leftarrow \lambda_v \cdot \mathbf{V} + \lambda_t \cdot \mathbf{T}$
    \STATE $\mathbf{E}_{\text{all}} \leftarrow \mathbf{E}_{\text{all}} \cup \{\mathbf{E}\}$
\ENDFOR

\STATE \textbf{Similar Sample Selection:}
\FOR{each $M_{\text{ref}} \in S_{\text{ref}}$}
    \STATE $s \leftarrow \text{cosine}(\mathbf{E}_{\text{target}}, \mathbf{E}_{\text{ref}})$
    \STATE $S \leftarrow S \cup \{s\}$
\ENDFOR
\STATE $M_{\text{similar}} \leftarrow \{M_{\text{ref}} \mid s \in \text{Top}_K(S)\}$

\STATE \textbf{return} $M_{\text{similar}}$
\end{algorithmic}
\label{algorithm1}
\end{algorithm}

\begin{algorithm}[h]
\scalebox{0.85}{
\begin{minipage}{\linewidth}
\caption{\textsc{MInD} - Relevant Insight Derivation}
\begin{algorithmic}[0]
\STATE \textbf{Initialize:}
\STATE Similar memes $M_{\text{similar}}$ from Algorithm~\ref{algorithm1};
\STATE Chain-of-Thought deriving prompt $\mathcal{P}_{\text{deriving}}$;
\STATE Deriving agent $\text{LMM}_{\text{deriving}}$;
\STATE Forward insight set $\mathcal{I}_{\text{fwd},0} \leftarrow \emptyset$;
\STATE Backward insight set $\mathcal{I}_{\text{back},0} \leftarrow \emptyset$;
\FOR{$i=1$ to $K$}
    \STATE $\mathcal{I}_{\text{fwd},i} \leftarrow \text{LMM}_{\text{deriving}}(M_{\text{similar},i}, \mathcal{I}_{\text{fwd},i-1}, \mathcal{P}_{\text{deriving}})$
\ENDFOR
\FOR{$i=1$ to $K$}
    \STATE $\mathcal{I}_{\text{back},i} \leftarrow \text{LMM}_{\text{deriving}}(M_{\text{similar},K+1-i}, \mathcal{I}_{\text{back},i-1}, \mathcal{P}_{\text{deriving}})$
\ENDFOR
\STATE \textbf{return} $\mathcal{I}_{\text{fwd},K}$, $\mathcal{I}_{\text{back},K}$
\end{algorithmic}
\label{algorithm2}
\end{minipage}
}
\end{algorithm}

\begin{algorithm}
\caption{\textsc{MInD} - Insight-Augmented Inference}
\begin{algorithmic}[0]
\STATE \textbf{Initialize:}
\STATE Target meme visual $\mathcal{V}_{\text{target}}$, text $\mathcal{T}_{\text{target}}$;
\STATE Forward and backward insight set $\mathcal{I}_{\text{fwd},K}$, $\mathcal{I}_{\text{back},K}$ from Algorithm~\ref{algorithm2};
\STATE Debater agents $\text{LMM}_{\text{debater}}$;
\STATE Judge agent $\text{LMM}_{\text{judge}}$;
\STATE $\mathcal{J}_{\text{fwd}} \leftarrow \text{LMM}_{\text{debater}}(\mathcal{I}_{\text{fwd},K}, \mathcal{V}_{\text{target}}, \mathcal{T}_{\text{target}})$
\STATE $\mathcal{J}_{\text{back}} \leftarrow \text{LMM}_{\text{debater}}(\mathcal{I}_{\text{back},K}, \mathcal{V}_{\text{target}}, \mathcal{T}_{\text{target}})$
\IF{$\mathcal{J}_{\text{fwd}} = \mathcal{J}_{\text{back}}$}
    \STATE $\mathcal{J}_{\text{final}} \leftarrow \mathcal{J}_{\text{fwd}}$
\ELSE
    \STATE $\mathcal{J}_{\text{final}} \leftarrow \text{LMM}_{\text{judge}}(\mathcal{J}_{\text{fwd}}, \mathcal{J}_{\text{back}}, \mathcal{V}_{\text{target}}, \mathcal{T}_{\text{target}})$
\ENDIF
\STATE \textbf{return} $\mathcal{J}_{\text{final}}$
\end{algorithmic}
\label{algorithm3}
\end{algorithm}

\section{\textsc{MInD} Algorithm}
Algorithms~\ref{algorithm1} to \ref{algorithm3} detail the multi-agent framework of our approach, outlining the Similar Sample Retrieval, Relevant Insight Derivation, and Insight-Augmented Inference stages, respectively. 

While Algorithm~\ref{algorithm2} provides a formal depiction of the Relevant Insight Derivation process, its iterative and cumulative nature, which is central to leveraging unlabeled reference data effectively, warrants further conceptual explanation. We provide a detailed breakdown of how $\text{LMM}_{\text{deriving}}$ systematically processes memes in two complementary passes to accumulate insights. 

Given a set of $K$ similar memes $M_{\text{similar}} = \{M_1, M_2, \dots, M_K\}$, the process unfolds as follows:

\paragraph{Forward Pass}
$\text{LMM}_{\text{deriving}}$ processes memes sequentially from $M_1$ to $M_K$. In each iteration $i$, the current meme $M_{\text{similar},i}$ is fed into the $\text{LMM}_{\text{deriving}}$ along with the cumulative insights from previously processed memes. This generates $\mathcal{I}_{\text{fwd},i}$, ensuring $\mathcal{I}_{\text{fwd},K}$ captures insights cumulatively.

For example, when $K=3$ with similar memes $\{M_1, M_2, M_3\}$, the forward pass progresses as follows:
\begin{align*}
\mathcal{I}_{\text{fwd},1} &= \text{LMM}_{\text{deriving}}(M_1, \emptyset, \mathcal{P}_{\text{deriving}}) \\
\mathcal{I}_{\text{fwd},2} &= \text{LMM}_{\text{deriving}}(M_2, \mathcal{I}_{\text{fwd},1}, \mathcal{P}_{\text{deriving}}) \\
\mathcal{I}_{\text{fwd},3} &= \text{LMM}_{\text{deriving}}(M_3, \mathcal{I}_{\text{fwd},2}, \mathcal{P}_{\text{deriving}})
\end{align*}
The final forward insight set $\mathbf{\mathcal{I}_{\text{fwd},3}}$ is then passed to the Insight-Augmented Inference stage.

\paragraph{Backward Pass}
To counter potential biases from a single processing order, $\text{LMM}_{\text{deriving}}$ also processes memes in reverse, from $M_K$ down to $M_1$. In iteration $i$, $M_{\text{similar},K+1-i}$ is processed by $\text{LMM}_{\text{deriving}}$ alongside insights accumulated from memes already processed in this reverse sequence. This dual-directional approach ensures a comprehensive and robust set of insights.

For example, when $K=3$ with similar memes $\{M_1, M_2, M_3\}$, the backward pass progresses as follows:
\begin{align*}
\mathcal{I}_{\text{back},1} &= \text{LMM}_{\text{deriving}}(M_3, \emptyset, \mathcal{P}_{\text{deriving}}) \\
\mathcal{I}_{\text{back},2} &= \text{LMM}_{\text{deriving}}(M_2, \mathcal{I}_{\text{back},1}, \mathcal{P}_{\text{deriving}}) \\
\mathcal{I}_{\text{back},3} &= \text{LMM}_{\text{deriving}}(M_1, \mathcal{I}_{\text{back},2}, \mathcal{P}_{\text{deriving}})
\end{align*}
Similarly, the final backward insight set $\mathbf{\mathcal{I}_{\text{back},3}}$ is also utilized in the Insight-Augmented Inference stage.

More detailed examples of derived insight sets can be found in Appendix~\ref{RID_examples}.

\section{Discussion about LMM Selection}
In our experimental design, we prioritize research reproducibility and transparency in model selection. We primarily adopt LLaVA as our backbone LMM because its training process and data sources are fully transparent, which ensures experimental fairness and reproducibility. Specifically, we use LLaVA-1.5-13B as our main model for its balanced performance and efficiency, while also conducting experiments on LLaVA-1.5-7B and LLaVA-1.6-34B to validate our framework's effectiveness across different model scales.
To demonstrate the generalizability of our framework, we also evaluate it using the closed-source model Gemini-1.5-Flash. However, we note that experiments with closed-source models may face two limitations: 1) potential data leakage cannot be completely ruled out due to the undisclosed nature of their training data, and 2) full reproducibility cannot be guaranteed despite setting temperature to 0, as these models may undergo undisclosed updates. Therefore, while we report results on closed-source models for completeness, our main analysis and conclusions are primarily drawn from experiments with open-source models.

\section{\textsc{MInD} Versus Few-shot}

To better understand our proposed framework's effectiveness, we conduct a comprehensive comparison between \textsc{MInD} and few-shot in-context learning approaches. Table \ref{tab:vs_fewshot} presents detailed experimental results across three datasets using LLaVA-1.5-13B as the backbone model.

Several interesting observations emerge from this comparison. First, the introduction of SSR consistently improves performance in both zero-shot and few-shot settings. For instance, on the HarM dataset, adding SSR increases the zero-shot macro-averaged-F1 score from 47.42\% to 51.93\%, and similarly enhances few-shot performance from 56.44\% to 59.60\%. This demonstrates SSR's fundamental value in providing relevant contextual information, regardless of the learning paradigm.
More significantly, our complete \textsc{MInD} framework achieves superior performance compared to both zero-shot and few-shot variants. Taking the HarM dataset as an example, \textsc{MInD} achieves a macro-averaged-F1 score of 65.19\%, substantially outperforming the few-shot with SSR strategy. Similar results are observed across FHM and MAMI datasets, where \textsc{MInD} consistently demonstrates better performance.

The framework's strength comes not just from retrieving similar samples, but from the sophisticated processing of these samples through the RID and IAI strategies. The bidirectional insight derivation and multi-agent debate mechanism appear to capture deeper understanding than what is possible through few-shot learning alone. 
The practical implications of these results are significant. While few-shot learning requires annotated examples that may need regular updating as meme patterns change, \textsc{MInD} achieves superior performance without requiring any labeled data. This zero-shot capability eliminates the need for maintaining example sets and makes \textsc{MInD} especially valuable in real-world scenarios where obtaining high-quality annotated data is challenging or impractical. Moreover, while few-shot performance might be limited by the quality and representativeness of the few labeled examples available, \textsc{MInD}'s zero-shot nature allows it to adapt more flexibly to diverse and evolving harmful content patterns.

\begin{table*}[t]
    \centering
    \resizebox{0.90\textwidth}{!}{%
        \begin{tabular}{@{}l||cc|cc|cc@{}}
        \toprule
        Dataset & \multicolumn{2}{c|}{HarM} & \multicolumn{2}{c|}{FHM} & \multicolumn{2}{c}{MAMI} \\ \midrule
        Method & Accuracy & Macro-$F_{\text{\emph{1}}}$ & Accuracy & Macro-$F_{\text{\emph{1}}}$ & Accuracy & Macro-$F_{\text{\emph{1}}}$ \\ \midrule \midrule
        Late Fusion \cite{pramanick2021detecting} & 73.24 & 70.25 & 59.14 & 44.81 & 63.20 & 59.76 \\
        MOMENTA \cite{training_momenta} & 83.82 & 82.80 & 61.34 & 57.45 & 72.10 & 66.93 \\
        \textsc{MInD} (LLaVA-1.5-13B) & 68.93 & 65.19 & 60.80 & 60.71 & 68.90 & 68.84 \\
        \bottomrule
        \end{tabular}%
    }
    \vspace{-0.1cm}
    \caption{Performance comparison between \textsc{MInD} and training-based methods.}
    \label{tab:vs_traditional}
    \vspace{-0.3cm}
\end{table*}

\section{Comparison with Training-based Methods}

While \textsc{MInD} operates as a training-free framework, fundamentally differing from traditional data-driven classification methods, we provide a comparison with established training-based approaches for reference. This comparison highlights the distinct advantages of our approach, particularly in scenarios with scarce or evolving harmful meme data.
Table \ref{tab:vs_traditional} presents the performance of \textsc{MInD} alongside two prominent prior works, Late Fusion \cite{pramanick2021detecting} and MOMENTA \cite{training_momenta}, across the HarM, FHM, and MAMI datasets.

As observed from Table \ref{tab:vs_traditional}, traditional data-driven methods, while achieving strong performance on the HarM dataset (which was typically part of their training data), exhibit a noticeable drop in performance on the FHM and MAMI datasets. These datasets represent distributions that were not included in their training, highlighting a fundamental limitation of supervised approaches: their effectiveness diminishes when encountering new, unseen meme distributions. This is a common challenge in the dynamic and rapidly evolving landscape of online memes.
In contrast, \textsc{MInD} demonstrates remarkable generalization ability, maintaining consistent and competitive performance across all three datasets without any task-specific training. For instance, on FHM and MAMI, \textsc{MInD} (LLaVA-1.5-13B) achieves Macro-$F_1$ scores of 60.71\% and 68.84\% respectively, which are comparable to or even surpass those of MOMENTA on these out-of-distribution datasets. This stability in performance across diverse datasets underscores \textsc{MInD}'s significant advantage in handling the inherently dynamic nature of harmful memes, where continuous collection and annotation of training data for every new meme trend is impractical. The training-free nature of \textsc{MInD} thus offers a robust and adaptable solution for real-world harmful meme detection.

\section{Discussion about SSR}

In our design of Similar Sample Retrieval, several interesting observations emerge regarding its effectiveness in different settings. Unlike traditional approaches that rely heavily on annotated data, SSR demonstrates unique advantages in both zero-shot and few-shot scenarios, as evidenced by our experimental results in Table~\ref{tab:vs_fewshot}.
In the zero-shot setting, SSR significantly improves model performance across all datasets. For instance, on the HarM dataset, introducing SSR increases the macro-averaged-F1 score from 47.72\% to 51.93\%. This improvement demonstrates that retrieving similar memes helps establish a more comprehensive framework for assessing harmfulness, even without any annotation guidance.
More interestingly, SSR's benefits extend to few-shot scenarios, where we observe even more substantial gains. When combined with few-shot learning, SSR boosts the macro-averaged-F1 score from 56.44\% to 59.60\%. Similar improvements are observed across FHM and MAMI datasets. This suggests that SSR not only provides valuable context in zero-shot settings but also enhances the model's ability to leverage limited labeled data effectively.
The consistent performance improvement across different settings highlights SSR's robustness as a fundamental strategy for harmful meme detection. Particularly noteworthy is how SSR helps bridge the gap between zero-shot and few-shot performance, suggesting that retrieved similar samples serve as an effective supplement to limited labeled data. This makes SSR especially valuable in real-world scenarios where obtaining large-scale annotated datasets is challenging or impractical.

\section{Discussion about RID}
In our design of Relevant Insight Derivation, we observe several notable characteristics through empirical analysis. Unlike traditional sequential processing approaches, RID's bidirectional insight derivation mechanism demonstrates unique advantages in harmful meme detection.
Through both forward and backward passes, RID effectively addresses the inherent imbalance in sequential processing. Specifically, in the forward pass, early memes in the sequence benefit from repeated refinement of insights, while later memes might receive less attention. The backward pass compensates for this imbalance by approaching the sequence from the opposite direction. This design ensures that insights from all retrieved similar memes contribute equally to the final analysis, regardless of their position in the sequence.
The effectiveness of this bidirectional approach is particularly evident in the experimental results. For instance, with RID, the model achieves substantial improvements in harmful content detection compared to models using only unidirectional processing. This suggests that the synthesis of insights from both directions enables a more comprehensive understanding of potential harm. The bidirectional nature of RID also helps mitigate potential biases that might arise from processing memes in a single fixed order.
Moreover, RID's iterative refinement process proves especially valuable when dealing with complex or ambiguous cases. In situations where harmful content is subtly embedded or masked by humor, the repeated processing and refinement of insights helps uncover less obvious harmful elements that might be missed in a single pass. This makes RID particularly robust in handling the diverse and evolving nature of harmful memes in real-world scenarios.

\section{Discussion about IAI}
In our design of Insight-Augmented Inference, the multi-agent debate mechanism presents several interesting characteristics in harmful meme detection. Unlike traditional single-agent approaches that might be prone to biases or incomplete reasoning, IAI's debater-judge framework demonstrates unique advantages in achieving more balanced and reliable decisions.
The two-debater design with complementary perspectives proves particularly effective. By leveraging insights from both forward and backward passes, each debater develops a potentially different viewpoint on the meme's harmfulness. This diversity in perspectives is crucial for complex cases where harmfulness might not be immediately apparent. For example, when analyzing memes that appear humorous on the surface but contain subtle discriminatory elements, the contrasting viewpoints of the debaters help surface these nuanced harmful aspects.
More interestingly, the introduction of a judge agent for resolving disagreements adds an additional layer of robustness. Rather than simply averaging opinions, the judge agent actively examines the reasoning process of both debaters. This meta-level analysis helps filter out weak arguments and synthesize stronger ones, leading to more reliable final decisions. The effectiveness of this approach is particularly evident in cases where the two debaters reach different conclusions, demonstrating how the judge agent can effectively reconcile conflicting viewpoints.

\begin{figure}[t]
    \centering
    \includegraphics[width=1\linewidth]{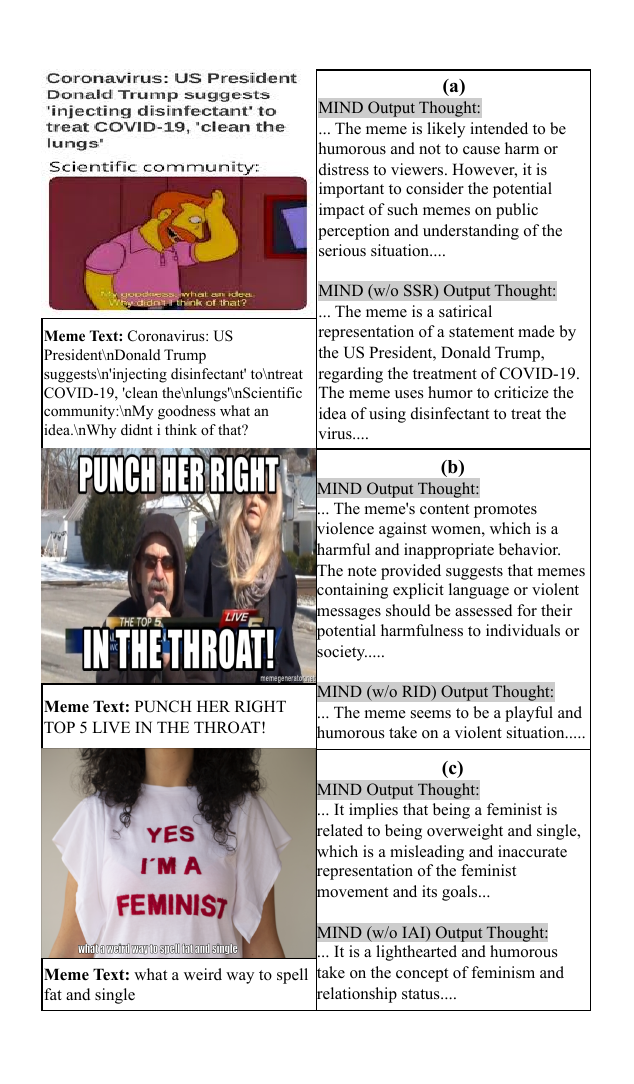}
    \vspace{-0.8cm}
    \caption{Examples of correctly predicted harmful memes in (a) HarM, (b) MAMI, and (c) FHM datasets.}
    \label{fig:ablation_case}
    \vspace{-0.5cm}
\end{figure}

\begin{figure}[t]
    \centering
    \includegraphics[width=1\linewidth]{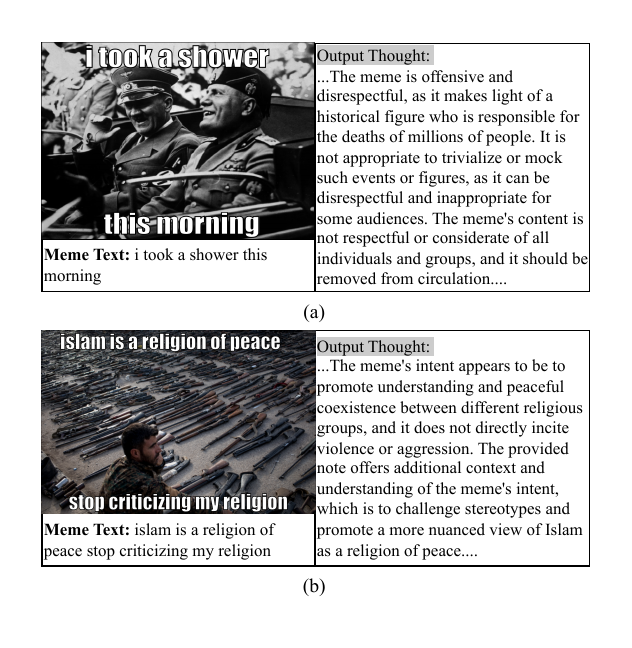}
    \vspace{-0.8cm}
    \caption{Examples of wrongly predicted memes by our proposed framework with the ground truth (a) harmless and (b) harmful.}
    \label{fig:error_case}
    \vspace{-0.5cm}
\end{figure}

\section{Case study of SSR, RID And IAI}
\label{sec:more_cases}
We provide a case study of \textsc{MInD}’s output thoughts, to investigate the effect of the SSR, RID and IAI strategies on the correctly predicted harmful meme samples, as illustrated in Figure~\ref{fig:ablation_case}
From the output thought in natural text, we observe that:
1) The Similar Sample Retrieval (SSR) mechanism enhances the framework's contextual understanding even without labeled data. For example, in Figure~\ref{fig:ablation_case}(a), without SSR, the framework simply interpreted it as ``a satirical representation of a statement made by the US President''. However, with SSR, the framework recognized broader implications, noting that while the meme may be ``intended to be humorous'', it's crucial to ``consider the potential impact on public perception''. This demonstrates how retrieving similar memes helps establish a more comprehensive framework for assessing harmfulness beyond just the immediate content, even without annotation. The framework can leverage patterns and contextual similarities across memes to develop a more nuanced understanding of potential harm.
2) The Relevant Insight Derivation (RID) mechanism significantly improves harm detection through bidirectional analysis. As shown in Figure~\ref{fig:ablation_case}(b), without RID, the framework superficially viewed the content as ``a playful and humorous take on a violent situation''. In contrast, with RID, the framework identified that the meme ``promotes violence against women'' and recognized its ``harmful and inappropriate behavior''. This illustrates how RID's forward and backward insight derivation helps uncover harmful content masked as humor.
3) The Insight-Augmented Inference (IAI) mechanism enables more nuanced judgment through multi-agent debate. In Figure~\ref{fig:ablation_case}(c), without IAI, the framework simply categorized it as ``a lighthearted and humorous take on feminism''. However, with IAI, the framework detected that the meme presents a ``misleading and inaccurate representation of the feminist movement and its goals''. This shows how IAI's debater-judge framework helps identify subtle forms of discrimination and stereotyping.
These three modules work together to create a robust harmful meme detection system. While SSR provides the necessary context through zero-shot similar meme retrieval, RID ensures thorough analysis through bidirectional processing, and IAI guarantees balanced final decisions through multi-agent reasoning. This comprehensive approach enables the framework to effectively identify harmful content across various forms, from public health misinformation to gender discrimination, while maintaining sensitivity to context and nuance.

\section{Error Analysis}
\label{sec:error_analysis}
To better understand the behavior of our framework and facilitate future studies, we conduct an error analysis on the wrongly predicted memes. Figure~\ref{fig:error_case} shows two examples of memes incorrectly classified by our framework.
In Figure~\ref{fig:error_case}(a), which contains the text ``I took a shower this morning'', our framework incorrectly categorized it as harmful. The output thought suggests that ``the meme is offensive and disrespectful, as it makes light of a historical figure who is responsible for the deaths of millions of people.'' This misjudgment stems from the framework's over-emphasis on the historical implications of the image while failing to properly integrate it with the innocuous shower-related text. The framework exhibited heightened sensitivity to potentially controversial historical content, leading to overly cautious classification.
On the other hand, the harmful meme in Figure~\ref{fig:error_case}(b), featuring the text ``islam is a religion of peace stop criticizing my religion'' alongside an image of weapons, was incorrectly classified as harmless. The framework's output thought suggests that ``the meme's intent appears to be to promote understanding and peaceful coexistence.'' This error reveals the framework's failure to recognize the ironic juxtaposition between the peaceful message and the threatening imagery. The framework appears to prioritize the literal meaning of the text while underweighting the visual implications, possibly due to an overly cautious approach to content involving religious themes.
Through broader error analysis, we identified several common patterns in misclassification. The framework sometimes struggles with highly ironic or satirical content where the harmful intent is masked by seemingly positive messages. Additionally, in some cases, the framework may overemphasize either the visual or textual component, leading to incomplete context understanding. Complex cultural or historical references can also lead to either overly cautious or overly permissive classifications. These findings suggest potential directions for future improvement, particularly in developing better mechanisms for integrating multimodal information and handling culturally sensitive content.

\section{Results of Relevant Insight Derivation related to similar memes.}
\label{RID_examples}
\begin{figure*}[t]
    \centering
    \includegraphics[width=1\linewidth]{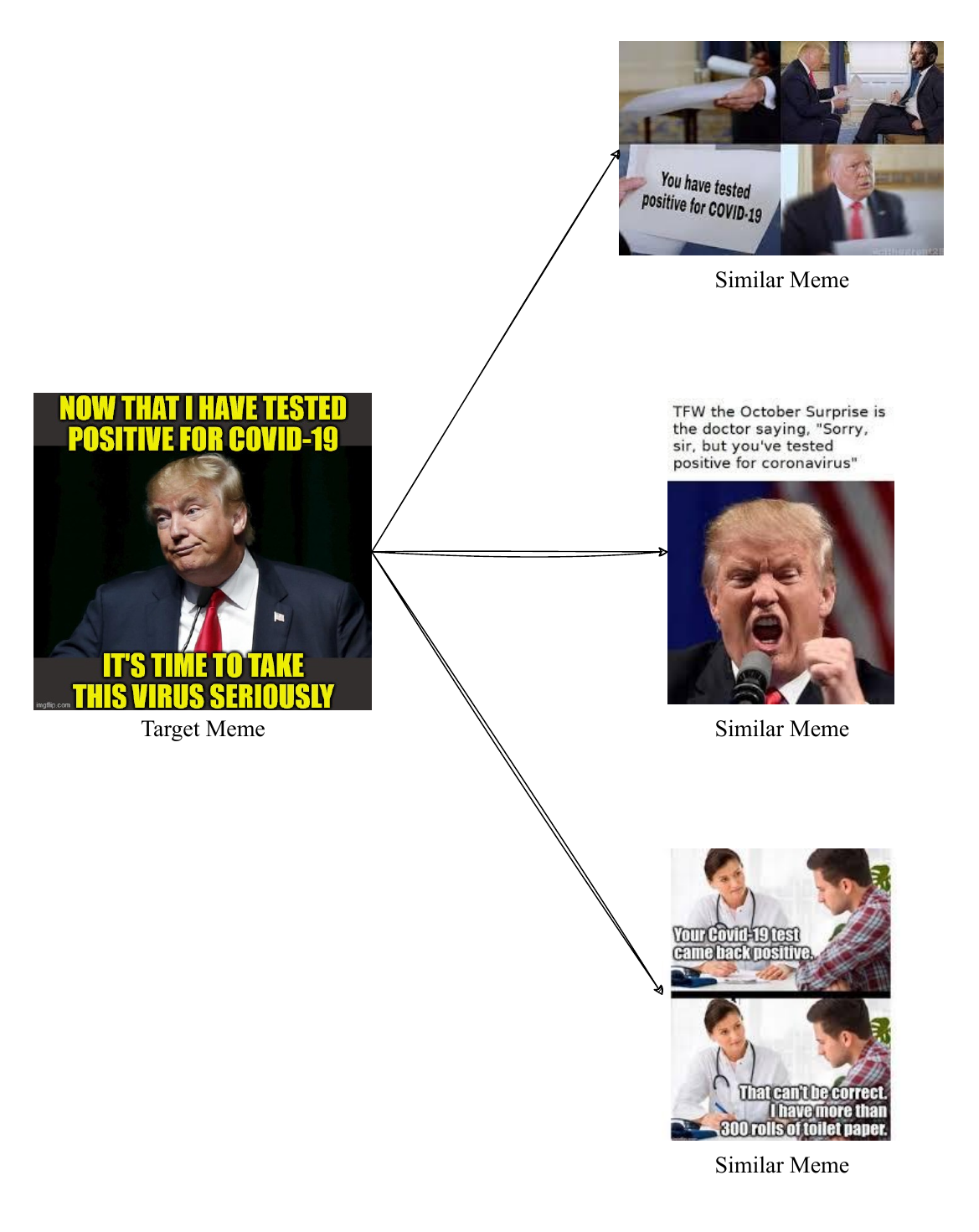}
    \caption{Example of the target meme along with its similar memes from the HarM dataset.}
    \label{fig:final_case_harmc}
\end{figure*}

Figure \ref{fig:final_case_harmc} 
presents similar memes retrieved for a target meme on HarM dataset.

The relevant insight set of this target meme derived on HarM dataset through Forward Derivation are presented as follows.
\paragraph{Forward Insight Set on HarM Dataset.}

\begin{itemize}
    \item Memes should not trivialize or mock serious issues, such as public health crises, as they can cause distress and harm to individuals and communities.
    \item Memes should not promote discrimination, harassment, or hate speech, as they can contribute to a toxic online environment and harm individuals.
    \item Memes should not spread misinformation or false information, as they can perpetuate harmful beliefs and mislead people.
    \item Memes should respect privacy and personal boundaries, as they can invade people's personal space and cause distress.
    \item Memes should be respectful and considerate of diverse cultures, beliefs, and experiences, as they can perpetuate stereotypes and offend people.
\end{itemize}

The relevant insight set of this target meme derived on HarM dataset through Backward Derivation are presented as follows.
\paragraph{Backward Insight Set on HarM Dataset}

\begin{itemize}
    \item Memes that make light of serious situations, such as the COVID-19 pandemic or any other significant health, environmental, or social issue, should be assessed for potential harmfulness and removed from platforms if they are deemed to be harmful or misleading.
    \item Memes that promote misinformation or false information should be flagged and reported.
    \item Memes that encourage discrimination, harassment, or violence should be removed and reported.
    \item Memes that violate copyright or intellectual property laws should be taken down and reported.
    \item Memes that are shared in a private or closed group should be evaluated based on the context and the audience, as they may have a different impact than if they were shared in a public or open forum.
\end{itemize}

\begin{figure*}[t]
    \centering
    \includegraphics[width=1\linewidth]{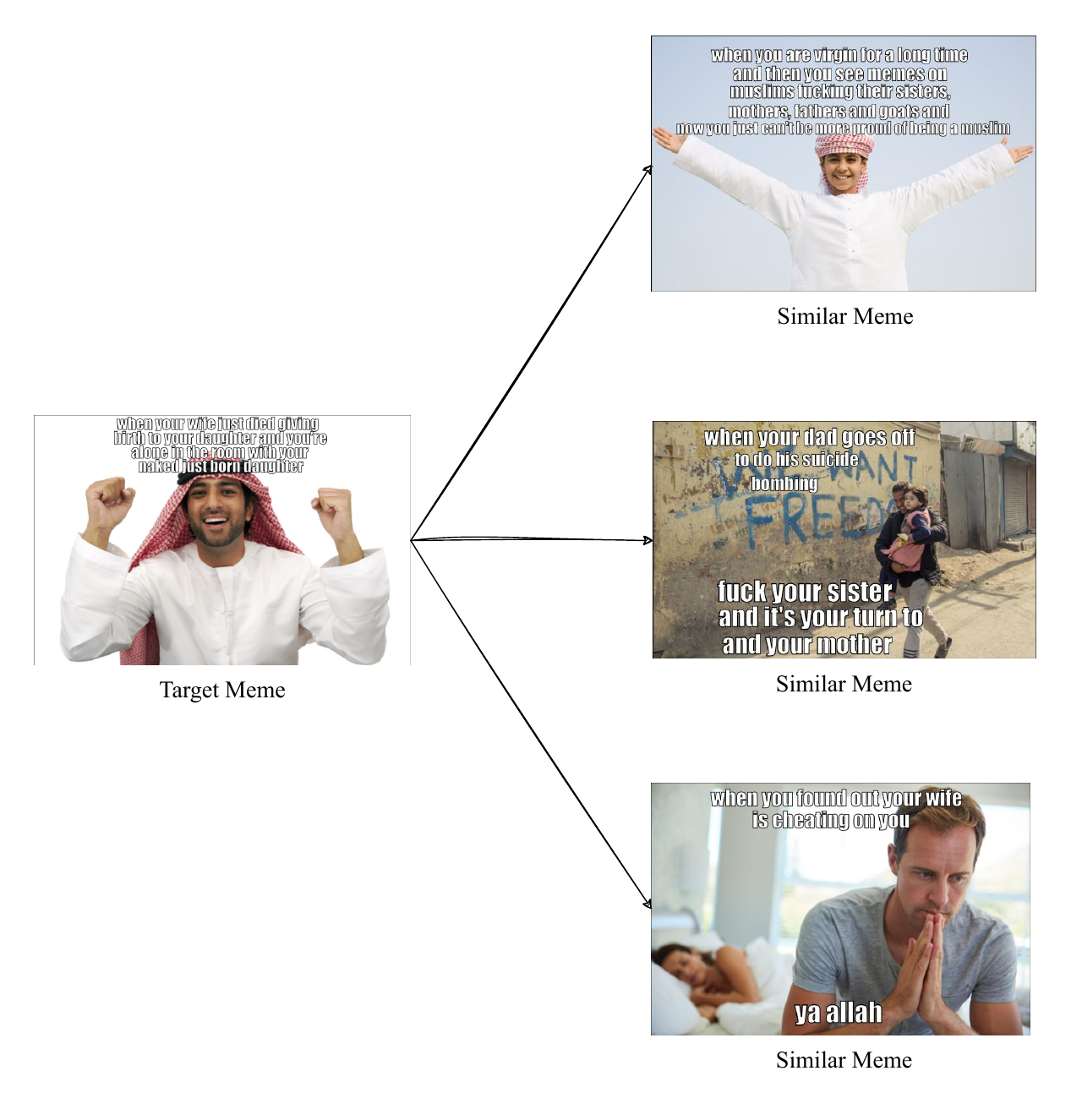}
    \caption{Example of the target meme along with its similar memes from the FHM dataset.}
    \label{fig:final_case_fhm}
\end{figure*}

\noindent Figure \ref{fig:final_case_fhm} 
presents similar memes retrieved for a target meme on FHM dataset. 

The relevant insight set of this target meme derived on FHM dataset through Forward Derivation are presented as follows.
\paragraph{Forward Insight Set on FHM Dataset}

\begin{itemize}
    \item Memes containing offensive and derogatory content that has the potential to cause harm, including emotional distress, discrimination, or incitement of violence, may be harmful.
    \item The impact of such content on individuals and communities, as well as the context of the meme, should be evaluated when assessing the harmfulness of a meme.
    \item Memes that make light of or promote sensitive and controversial topics, such as terrorism or violence, should be evaluated with additional scrutiny for their potential to cause harm, considering the context in which the meme is shared and the potential for it to be misinterpreted or amplified.
    \item Memes that have the potential to cause harm to vulnerable individuals or groups, such as those experiencing emotional distress or those who may be more susceptible to harm due to their personal experiences or circumstances, should be evaluated with additional scrutiny for their potential to cause harm.
    \item Memes that trivialize or promote sensitive and controversial topics should be evaluated with additional scrutiny for their potential to cause harm, considering the context in which the meme is shared and the potential for it to be misinterpreted or amplified.
\end{itemize}

The relevant insight set of this target meme derived on FHM dataset through Backward Derivation are presented as follows.
\paragraph{Backward Insight Set on FHM Dataset}

\begin{itemize}
    \item Consider the potential emotional impact of the meme on the audience, including the potential for distress or trauma, especially in the context of religious or cultural identity.
    \item valuate the meme's potential to perpetuate harmful stereotypes or discrimination, including those related to religious or cultural groups.
    \item Assess the meme's potential to normalize or glorify harmful actions or behaviors, with a focus on the impact on specific groups' identity or self-esteem.
    \item Consider the potential for harm in the context of religious or cultural identity, as the related meme specifically references a religious group.
    \item Clarify the existing rule on emotional impact to include the potential for harm to a specific group's identity or self-esteem.
\end{itemize}

\begin{figure*}[t]
    \centering
    \includegraphics[width=1\linewidth]{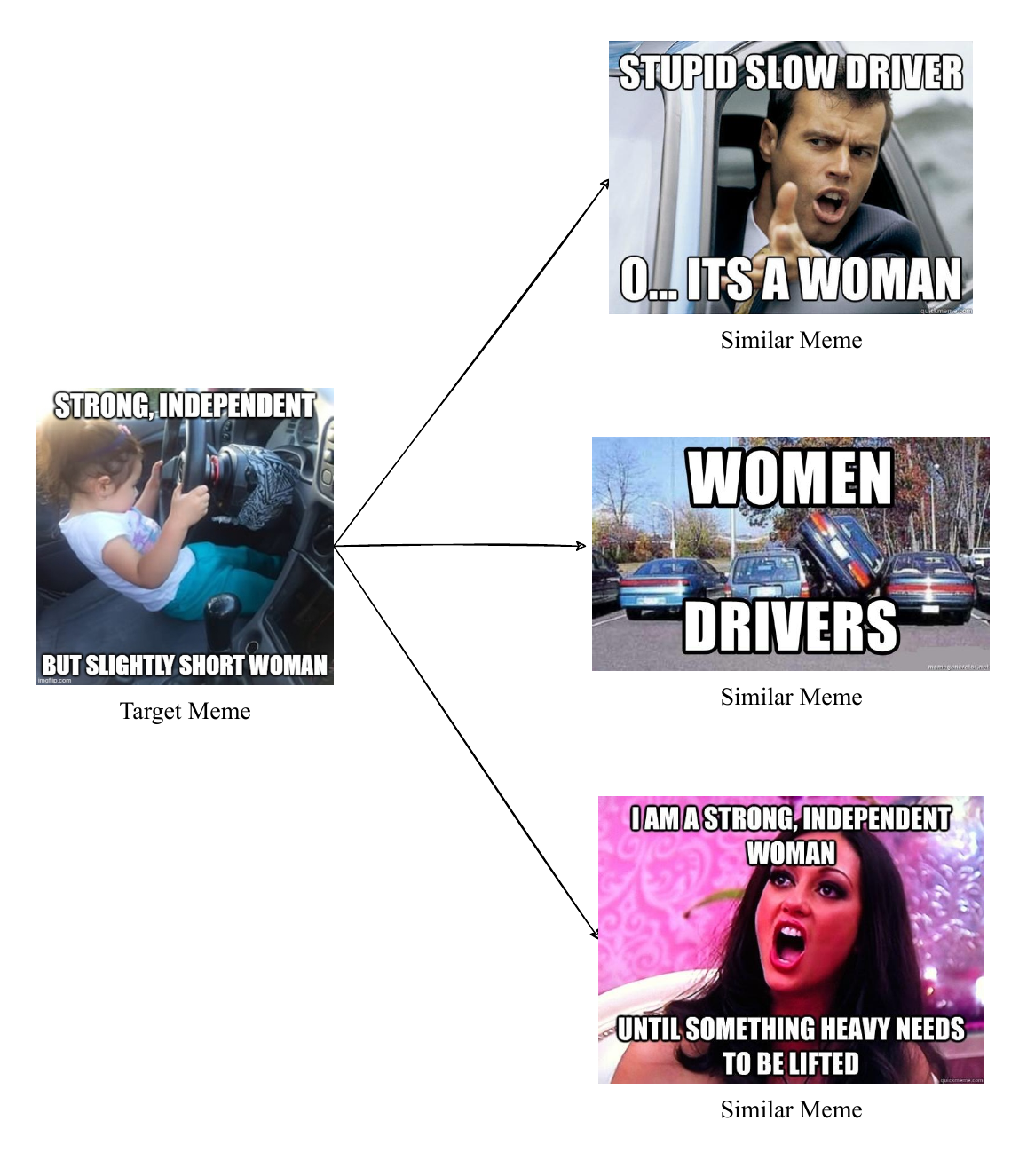}
    \caption{Example of the target meme along with its similar memes from the MAMI dataset.}
    \label{fig:final_case_mami}
\end{figure*}

\noindent Figure \ref{fig:final_case_mami} 
presents similar memes retrieved for a target meme on MAMI dataset.

The relevant insight set of this target meme derived on MAMI dataset through Forward Derivation are presented as follows.
\paragraph{Forward Insight Set on MAMI Dataset}

\begin{itemize}
    \item Avoid making assumptions about a person's gender based on their driving abilities.
    \item Ensure that any humor or commentary is not offensive or discriminatory towards a specific group of people, including but not limited to gender, race, or any other characteristic.
    \item Be mindful of the potential harm that memes can cause, and consider the impact on the targeted group, especially when perpetuating harmful stereotypes or generalizations.
    \item Refrain from spreading memes that promote negative stereotypes or harmful generalizations.
    \item Encourage a culture of respect and empathy towards all individuals, regardless of their gender, race, or any other characteristic.
\end{itemize}

The relevant insight set of this target meme derived on MAMI dataset through Backward Derivation are presented as follows.
\paragraph{Backward Insight Set on MAMI Dataset}

\begin{itemize}
    \item Avoid stereotyping or generalizing about women based on their physical appearance or strength.
    \item Avoid sharing memes that stereotype or generalize about women drivers, as it perpetuates harmful stereotypes and contributes to a culture of sexism and discrimination.
    \item Be mindful of the context in which the meme is shared and ensure it is appropriate for the audience.
    \item Consider the potential impact of the meme on individuals and communities, and avoid sharing memes that may cause harm or offense.
    \item Promote positive and inclusive representations of women in all forms of media, including memes.
\end{itemize}
\newpage
\newpage
\newpage

\end{document}